# Linearized subspace refinement framework to expose hidden accuracy in trained neural networks

Wenbo Cao[1,2,3], Weiwei Zhang[3,4,5,*]

[1] *Institute of AI for Industries, Chinese Academy of Sciences, Nanjing 211135, China*

[2] *Institute of Computing Technology, Chinese Academy of Sciences, Beijing 100190, China*

[3] *School of Aeronautics, Northwestern Polytechnical University, Xi'an 710072, China*

[4] *International Joint Institute of Artificial Intelligence on Fluid Mechanics, Northwestern Polytechnical University, Xi'an 710072, China*

[5] *National Key Laboratory of Aircraft Configuration Design, Xi'an 710072, China*

* *Correspondence to: aeroelastic@nwpu.edu.cn*

**Abstract.** Neural networks trained by gradient-based methods often exhibit optimization-induced accuracy plateaus in scientific machine learning tasks. We present Linearized Subspace Refinement (LSR), an architecture-agnostic post-training framework that exploits the local linearized model at a fixed trained state. By solving a reduced direct least-squares problem in a Jacobian-defined low-dimensional space, LSR computes a subspace-optimal linearized correction and yields a refined predictor with markedly improved accuracy. Across function approximation, data-driven operator learning, physics-informed operator fine-tuning, and noisy inverse problems, LSR shows that standard nonlinear training can remain far above this subspace-attainable error level. Similar accuracy plateaus persist even for the convex quadratic problem from local linearization when solved with standard iterative optimizers, identifying numerical ill-conditioning as a primary bottleneck. LSR frequently delivers order-of-magnitude error reductions, while the subspace rank provides an explicit capacity-control mechanism that balances correction strength, numerical stability, and noise sensitivity. Together, LSR exposes conditioning-limited attainable accuracy in trained-state linearized models and provides direct access to it.

## Introduction

Neural networks, owing to their expressive power and flexibility[1], are widely used as function approximators in supervised regression, operator learning, and physics-informed scientific computing[2-5]. In these applications, the objective is often not merely to obtain qualitatively correct predictions, but to approximate continuous solutions or operators with high numerical accuracy. For example, in operator modeling and partial differential equation (PDE) solving, the achievable error magnitude directly determines whether a learned model is suitable for scientific analysis or engineering deployment[2,3]. Consequently, this work addresses a central question: in these accuracy-critical settings, why does gradient-based training often stall at an accuracy plateau, and how can one systematically push accuracy beyond it under a fixed architecture and standard training procedure?

From the perspective of classical numerical analysis and linear algebra, ill-conditioning is widely recognized as a key factor limiting attainable accuracy. Even when a problem is linear or convex in theory, severe ill-conditioning can prevent iterative methods from reaching high-accuracy solutions within a practical computational budget. In contrast, when memory and computational cost permit, direct factorization-based solvers can reliably attain much higher accuracy by bypassing the slow error decay typical to iterative schemes[6,7]. This fundamental distinction underlies long-standing trade-offs between accuracy, stability, and efficiency in numerical computation, and provides a useful lens for re-examining accuracy barriers encountered in neural network models[8].

Despite these insights, modern neural network training remains overwhelmingly dominated by gradient-based iterative optimization[9,10]. In this paradigm, ill-conditioning manifests as slow convergence, training instability, and strong sensitivity to hyperparameter choices, motivating a wide range of increasingly sophisticated optimizers and acceleration strategies, including adaptive learning rates, momentum, quasi-second-order updates, and various forms of preconditioning[11-16]. In parallel, substantial effort has targeted improving accuracy in scientific machine learning through architectural inductive biases and problem-specific design choices, encompassing network architectures, sampling strategies, loss reweighting, regularization, curriculum learning, and multi-stage training[2,3,17-23]. While diverse in form, these approaches share a common feature: they pursue accuracy improvements almost exclusively by continuing gradient-based updates of the parameter vector $\boldsymbol{\theta}$.

Motivated by these considerations, we consider a complementary regime to standard training. Rather than asking how to further optimize the parameter vector through continued gradient-based training, we ask the following: after training appears to have converged and the parameters are fixed at $\boldsymbol{\theta}_0$, do directions exist in the locally linearized model (induced by the network Jacobian) that can still substantially reduce the loss? More fundamentally, can direct solvers bypass conditioning barriers that limit the accuracy reached by iterative optimization under practical budgets? Beyond providing a refinement route, this perspective offers a concrete and falsifiable probe of attainable accuracy in nonlinear neural networks. While optimality of the original nonconvex objective is generally uncertifiable at scale, the local linearization at $\boldsymbol{\theta}_0$ defines a convex quadratic least-squares problem whose subspace-attainable solution can be computed to high accuracy with direct solvers when feasible, thereby quantifying the locally accessible improvement left unused by standard training and diagnosing accuracy plateaus beyond nonconvexity alone.

Local linearization has also been extensively studied in the neural tangent kernel (NTK) literature[24-28], where tangent features around random initialization are shown to define rich function classes in wide-network or infinite-width regimes. These results clarify the expressive capacity and training dynamics of linearized neural networks. The present work addresses a different but complementary question. Rather than asking whether the tangent space is expressive in principle, we ask whether the accuracy available in the local linearized model of a finite trained network is numerically attainable under finite precision and finite computational budgets. This distinction is essential because a rich linearized function class does not necessarily imply a well-conditioned least-squares problem, nor does it guarantee that gradient-based optimization can stably reach the high-accuracy solution in practice. Moreover, our focus is on post-training refinement: LSR is applied after standard nonlinear training has reached an apparent accuracy plateau, and further improves accuracy by exploiting the Jacobian-induced response space at the trained state, rather than relying on tangent features at random initialization.

A key observation underlying this perspective is that many seemingly disparate learning tasks share a common objective structure. Supervised regression, data-driven operator learning, and physics-informed PDE solving can all be formulated as minimizing a residual norm,

$$\min_{\boldsymbol{\theta}} \frac{1}{2}\|\boldsymbol{f}(\boldsymbol{\theta})\|_2^2 \tag{1}$$

where $\boldsymbol{f}(\boldsymbol{\theta})$ denotes the full-training-set residual vector, constructed by evaluating the model prediction $\boldsymbol{q}(\boldsymbol{\theta})$ at all data samples, observation points, or collocation constraints used in training and assembling the corresponding task-specific residuals. For notational simplicity, we write $\boldsymbol{f}(\boldsymbol{\theta})$ rather than $\boldsymbol{f}(\boldsymbol{q}(\boldsymbol{x};\boldsymbol{\theta}))$ . The primary differences among these problems lie in the composition and interpretation of the residuals, rather than in the residual-minimization form itself. This structural universality provides a natural entry point for introducing a unified refinement framework grounded in classical direct solvers.

Linearizing the residual vector around a trained model leads to a classical least-squares problem involving the network Jacobian. However, the Jacobian is typically extremely high-dimensional and dense, rendering direct solves over the full parameter space computationally infeasible. To address this challenge, we introduce the Linearized Subspace Refinement (LSR) framework. LSR constructs a compact correction subspace using highly parallel Jacobian–vector products and computes a subspace-optimal refinement by directly solving a reduced linear residual problem within this subspace. LSR operates as an architecture-agnostic post-training procedure: it does not modify network architectures, loss formulations, or the original training process, yet it can expose substantial additional accuracy that is not exploited by gradient-based training alone under standard training procedures. The correction-subspace rank further provides an explicit capacity-control mechanism, balancing correction strength, numerical stability, and noise sensitivity. We demonstrate this rank-controlled post-training refinement across function approximation, data-driven operator learning, physics-informed operator fine-tuning, and noisy inverse problems.

## Results

### Method overview

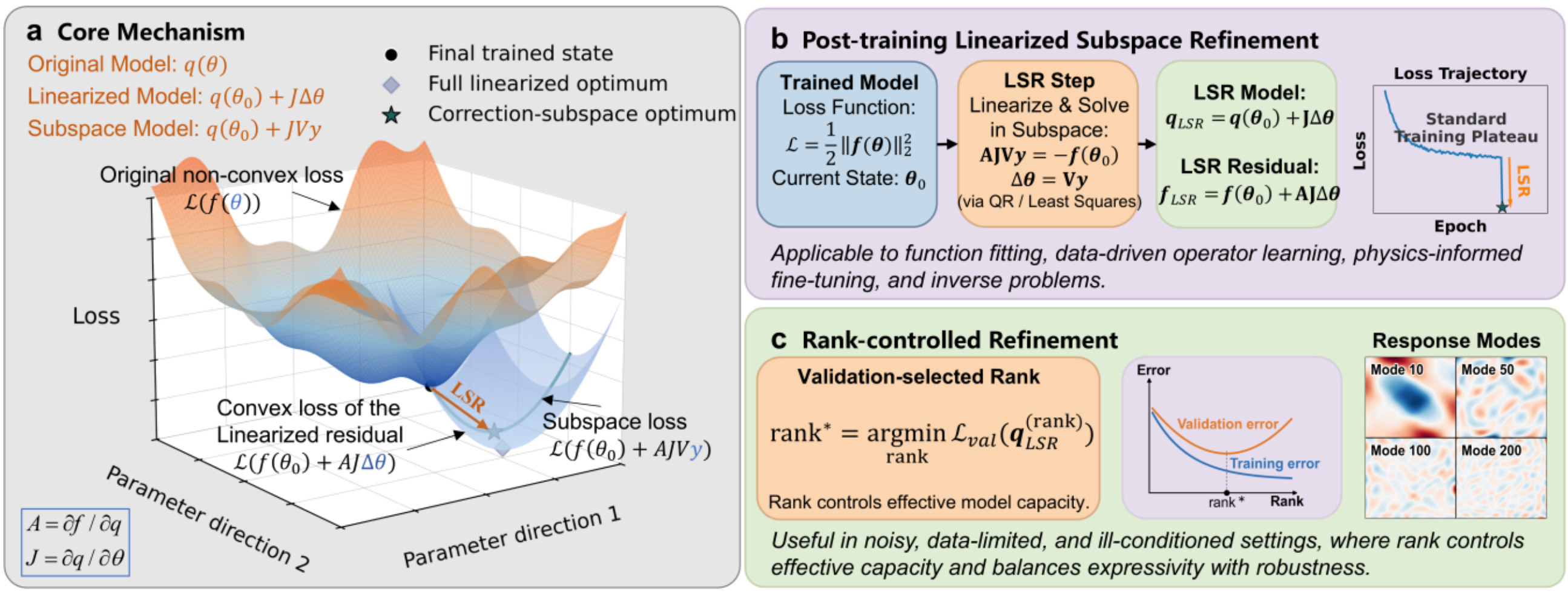

**Fig. 1. Overview of the Linearized Subspace Refinement (LSR) framework. a, Core mechanism.** Starting from a trained network at parameters $\boldsymbol{\theta}_0$, the residual vector is locally linearized, inducing a convex least-squares problem with respect to $\Delta\boldsymbol{\theta}$. By restricting $\Delta\boldsymbol{\theta}$ to a low-dimensional subspace, an explicit and computationally tractable direct linear solve yields the subspace-optimal correction. **b, Post-training LSR.** After training plateaus, LSR exposes additional attainable accuracy through a local linearized correction without changing the original model or training pipeline. **c, Rank-controlled refinement.** The LSR rank controls the amount of Jacobian-induced response space used for refinement and provides an explicit capacity-control mechanism. Lower ranks typically retain smoother and more robust response modes, whereas higher ranks introduce finer-scale modes that improve expressivity but may reduce robustness in noisy, data-limited, or ill-conditioned settings.

The core mechanism of LSR is illustrated in Fig. 1. Starting from a trained parameter vector $\boldsymbol{\theta}_0$, LSR locally linearizes the full-set residual vector with respect to a correction variable $\Delta\boldsymbol{\theta}$, which yields a convex linear least-squares model,

$$\min_{\Delta\boldsymbol{\theta}} \left\| \boldsymbol{f}(\boldsymbol{\theta}_0) + \mathbf{AJ}\Delta\boldsymbol{\theta} \right\|_2^2 \tag{2}$$

where $\mathbf{J} = \partial \boldsymbol{q} / \partial \boldsymbol{\theta}$ denotes the Jacobian of the network output over the training set with respect to parameters, and $\mathbf{A} = \partial \boldsymbol{f} / \partial \boldsymbol{q}$ represents the Jacobian of the residual assembly with respect to outputs. To make direct solvers computationally tractable, LSR restricts $\Delta\boldsymbol{\theta}$ to a low-dimensional subspace $\mathbf{V}$, yielding the optimal solution within this subspace at a manageable computational cost. We refer to $\mathbf{V}$ as the Jacobian-induced parameter-space correction subspace, and to its image under the output Jacobian $\mathbf{JV}$, as the corresponding Jacobian-induced response space. Importantly, the resulting $\Delta\boldsymbol{\theta}$ is not applied as a nonlinear parameter update. Instead, LSR evaluates the linearized solution at the same linearization point to define a refined predictor, improving accuracy without assuming global validity of the linear approximation. For any test input $x$, the refined predictor is defined as

$$q_{LSR}(x) = q(x;\boldsymbol{\theta}_0) + \mathbf{J}(x;\boldsymbol{\theta}_0)\Delta\boldsymbol{\theta} \tag{3}$$

For practical inference, this prediction requires only the Jacobian-vector product $\mathbf{J}(x;\boldsymbol{\theta}_0)\Delta\boldsymbol{\theta}$, not the full pointwise Jacobian. In our implementation, its measured cost in the tested example is approximately 2-4 times that of a standard forward evaluation (Supplementary Section 3). For brevity, we use $\mathbf{J}$ to denote Jacobians evaluated at the appropriate inputs, with the understanding that training-set and pointwise Jacobians are distinguished by context.

The correction-subspace rank controls how much of the Jacobian-induced response space is used for refinement. Since the correction basis is constructed by randomized singular value decomposition (RSVD) to approximate the leading right singular directions of the Jacobian, the basis directions are ordered by descending singular value. Because RSVD is randomized, bases computed independently at different ranks are not necessarily column-wise nested, but they approximate the same ordered hierarchy of leading right singular subspaces. A response mode is defined as the linearized output perturbation generated by activating one normalized basis direction. Specifically, for the $i$-th normalized basis vector $\boldsymbol{v}_i$, setting $\Delta\boldsymbol{\theta} = \boldsymbol{v}_i$ gives $\delta\boldsymbol{q}_i = \mathbf{J}\boldsymbol{v}_i$. As illustrated in Fig. 1c, leading modes typically correspond to smoother, larger-scale corrections, whereas higher-index modes can introduce finer and more oscillatory responses. Thus, increasing the rank progressively adds correction directions with greater expressive detail but also higher sensitivity to ill-conditioning or noise.

In the experiments below, each baseline model is first trained using its standard training procedure, and LSR is then applied to the resulting checkpoint. Depending on the purpose of the experiment, the correction-subspace rank is either scanned to analyze attainable accuracy and conditioning effects or selected by validation performance to control capacity under noise. All experiments reported in this paper are conducted on a single NVIDIA A100 GPU with 40 GB of memory.

**Function approximation**

We begin with a standard two-dimensional function approximation problem to assess the attainable accuracy of trained neural networks and to examine the behavior of LSR in the simplest supervised learning setting. The target function is $q(x, y) = \sin(\pi x)\sin(\pi y)$, defined on the square domain $[-1,1]\times[-1,1]$, and is approximated using a conventional feed-forward neural network. Network parameters are trained using the Adam optimizer with mini-batches until the test mean squared error (MSE) reaches a clear plateau. To isolate the numerical limits of gradient-based training, this experiment is conducted in an idealized supervised setting with dense, noise-free data and sufficient model capacity, so that representation error and data noise do not constitute the dominant sources of error. Full details of the network architecture and training procedure are provided in Supplementary Section 2.1.

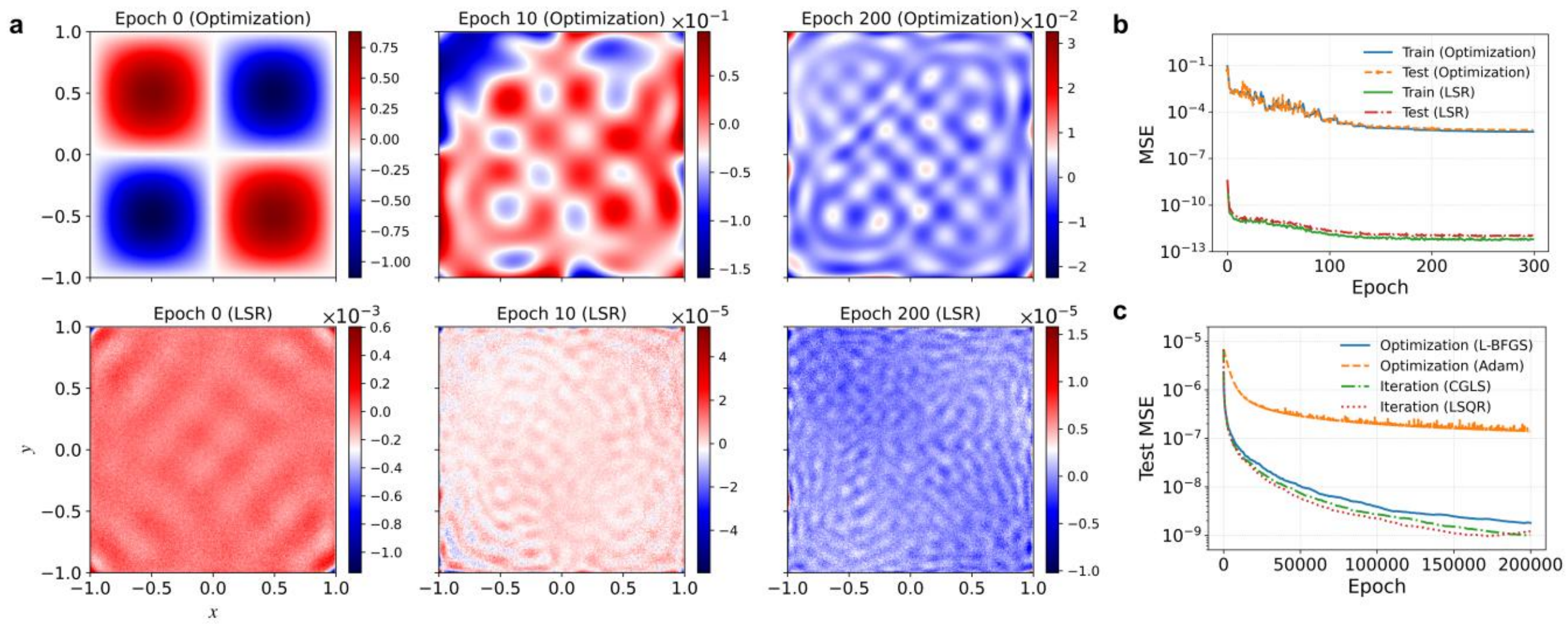

**Fig. 2. Accuracy enhancement and characteristic behavior of LSR in a two-dimensional function approximation problem. a,** Spatial distribution of the network prediction error at different training stages, together with the prediction error after applying LSR at the same parameter states. **b,** Evolution of the MSE during training. Curves labeled "LSR" indicate the error levels attained by applying LSR at the corresponding training epochs. **c,** Error convergence for the convex least-squares problem obtained from local linearization of the residual, solved using different optimization or iterative methods.

Figure 2a shows the spatial distribution of the prediction error at different training epochs (0, 10 and 200), together with the error obtained by applying LSR at the same parameter states. As training progresses, gradient-based optimization progressively suppresses low-frequency error components, while the remaining discrepancy becomes increasingly dominated by high-frequency oscillations, consistent with the well-known spectral bias of neural network training[29,30]. In contrast, LSR markedly reduces the error magnitude, leaving predominantly fine-scale oscillatory structures. This comparison indicates that, for this simple supervised task, standard training effectively captures the dominant low-frequency content of the target function, whereas the additional accuracy exposed by LSR primarily arises from correcting localized high-frequency errors.

The evolution of the training and test MSE during optimization is shown in Fig. 2b. The curves labeled "LSR" report the MSE level obtained by applying LSR to the network parameters at the corresponding training epoch. Although gradient-based training rapidly reduces the error at early stages, progress slows substantially once a plateau is reached. By contrast, LSR consistently produces order-of-magnitude MSE reductions throughout training. When applied in the plateau regime, LSR further reduces the test MSE from approximately $10^{-6}$ to nearly $10^{-12}$. In this setting, the wall-clock cost of LSR is negligible relative to training, requiring less than one second in our implementation.

To further pinpoint the origin of this accuracy limitation, Fig. 2c isolates the local linearized residual model at the trained network state $\boldsymbol{\theta}_0$ and compares how different iterative algorithms solve the same full-space convex least-squares problem, including first-order Adam, quasi-Newton Limited-memory Broyden–Fletcher–Goldfarb–Shanno (L-BFGS), as well as classical Krylov solvers for least-squares problems, namely Conjugate Gradient for Least Squares (CGLS) and LSQR. It is important to note that the LSR solution, obtained by solving a reduced least-squares problem within a low-dimensional correction subspace, yields a conservative estimate of the achievable error reduction, since the reduced subspace is a strict subset of the full parameter space. Despite the fact that the objective is convex, the accuracy achieved by these iterative methods within a practical iteration budget remains significantly lower than that achieved by the direct solve of the reduced problem. Among the iterative approaches, L-BFGS performs noticeably better than first-order Adam, reflecting partial access to curvature information, while CGLS and LSQR achieve accuracy comparable to L-BFGS. Additional results under different optimizer hyperparameter settings are reported in Supplementary Fig. 3, demonstrating that each optimization method converges to a similar accuracy plateau across a wide hyperparameter range, which remains substantially above the LSR direct-solve error even after up to $10^6$ iterations.

These results show that the accuracy plateau is already visible in the local linearized least-squares problem, where nonconvexity has been removed. The gap between iterative solvers and the reduced direct solve therefore points to numerical conditioning as a practical bottleneck for attainable accuracy. It is worth emphasizing that the LSR correction is not applied as a nonlinear parameter update. Although the linearized least-squares problem admits a solution with substantially lower residual, directly reinjecting the corresponding correction direction into the nonlinear network generally does not reproduce the refined prediction and may even degrade performance, as shown in Supplementary Fig. 13.

**Data-driven operator learning**

We next examine the behavior of LSR in a data-driven operator learning setting based on the one-dimensional Burgers equation. The task is to learn the operator mapping from the initial condition $q(x,t=0)$ to the solution at a later time $q(x,t=1)$, where the initial conditions are sampled from a Gaussian random field. Two representative neural operator architectures, DeepONet[2] and MultiONet[31], are considered, together with three different activation functions (ReLU, Tanh, and a custom Tanh–Sin activation), in order to assess the robustness of LSR across network architectures and activation-induced nonlinear representations. All models are trained using standard supervised learning to obtain baseline solutions, after which LSR is applied as a post-training refinement.

Figure 3a reports aggregate statistics over six operator-learning configurations, combining different architectures and activation functions. Unless otherwise specified, all reported test errors are measured using the relative $L_2$ norm. Across all settings, LSR consistently reduces the test error compared with the baseline predictions, with the paired distributions revealing systematic accuracy gains across individual test instances rather than isolated improvements. Importantly, instances with lower baseline errors tend to exhibit smaller post-LSR errors, indicating that LSR amplifies—rather than replaces—the benefits of effective nonlinear training. These results demonstrate that the benefits of LSR are robust across architectures and activations, and are not restricted to a small subset of favorable cases.

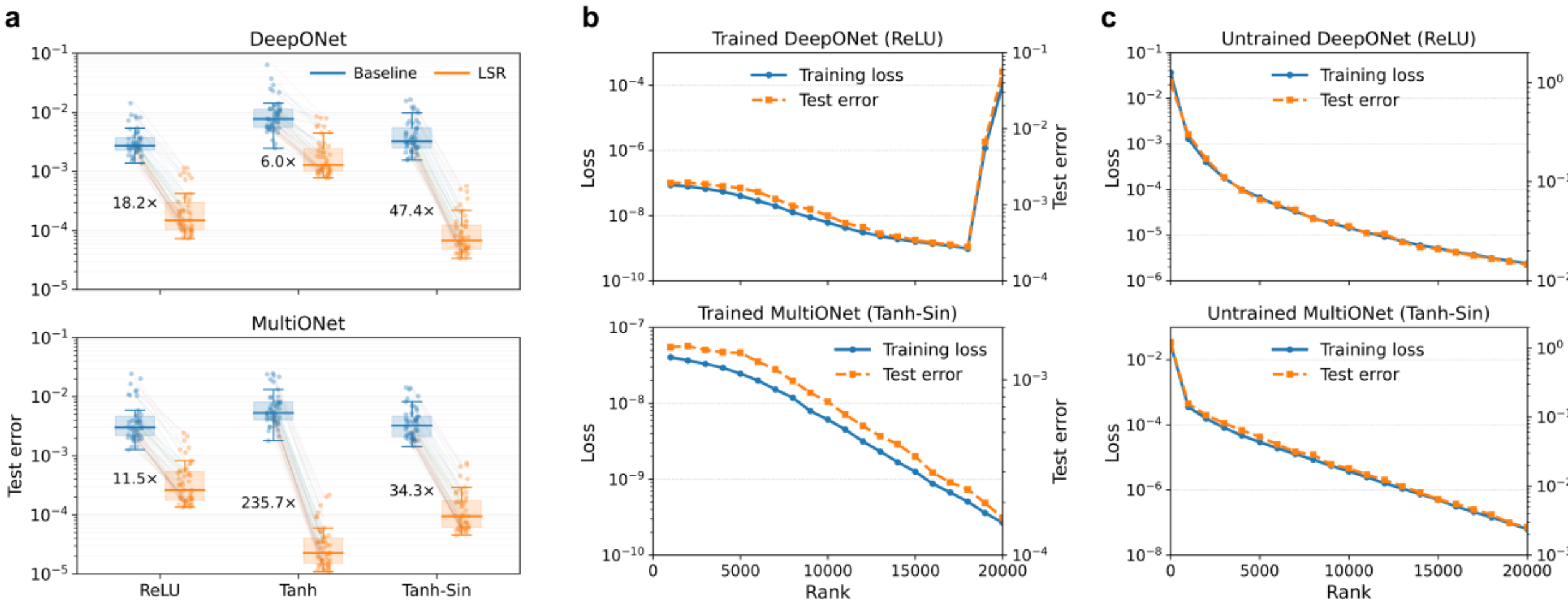

**Fig. 3. Accuracy gains from linearized subspace refinement in data-driven operator learning. a,** Aggregate error statistics over six data-driven operator-learning configurations. Each point corresponds to one of 50 test instances, with paired distributions comparing baseline predictions and their LSR counterparts. Reported numbers indicate the median error reduction factors. **b,** Rank-dependent training loss and test error for representative trained DeepONet and MultiONet models. Increasing the rank allows LSR to expose lower attainable error levels over a broad rank range. **c,** Corresponding rank-dependent behavior for randomly initialized networks, included as a reference to demonstrate that the Jacobian-induced response space already possesses substantial representational capacity even prior to training.

Figures 3b and 3c further elucidate how these gains depend on the rank of the correction subspace. For representative trained DeepONet and MultiONet models (Fig. 3b), increasing the rank leads to a pronounced and sustained reduction in both training loss and test error. This monotonic trend over a wide range of ranks indicates that the accuracy improvements provided by LSR do not arise from correcting only a few dominant parameter directions. Instead, they reflect the progressive exploitation of additional correction directions admitted by the linearized model as the subspace expands. The corresponding results for randomly initialized networks (Fig. 3c) reveal that, even prior to training, the Jacobian-induced response space already possesses substantial representational capacity. This behavior is consistent with the NTK perspective, which shows that tangent features around random initialization can define rich function classes in wide-network regimes. In these random-initialization experiments, all linear layers use the default PyTorch nn.Linear initialization, namely Kaiming-uniform initialization for the weights and uniform initialization for the biases.

At the same time, the comparison between Fig. 3b, c highlights the role of training in shaping the finite-network Jacobian-induced response space. Trained models allow LSR to reach substantially lower attainable error floors than randomly initialized models. To separate the effect of the improved base prediction from the effect of the Jacobian-induced response space, we further performed a controlled comparison using the same trained base prediction and residual. Specifically, we constructed LSR corrections using either the Jacobian at random initialization or the Jacobian at the trained state: $q = q(\boldsymbol{\theta}_{\text{train}}) + \mathbf{J}(\boldsymbol{\theta}_{\text{init}})\Delta\boldsymbol{\theta}$ and $q = q(\boldsymbol{\theta}_{\text{train}}) + \mathbf{J}(\boldsymbol{\theta}_{\text{train}})\Delta\boldsymbol{\theta}$. As shown in Supplementary Fig. 6, the correction based on the trained Jacobian consistently achieves lower loss and test error than that based on the randomly initialized Jacobian. This result indicates that the trained-state advantage is not only due to a better base prediction; training also reshapes the Jacobian-induced response space, making it more effective for correcting the current residual. This behavior is distinct from the fixed-kernel picture associated with the idealized infinite-width NTK regime and supports our focus on finite trained networks.

At sufficiently large ranks, the monotonic improvement with increasing subspace dimension may eventually deteriorate, with the loss and error exhibiting abrupt increases. In exact arithmetic, enlarging the correction space should not increase the optimal residual; therefore, this behavior reflects a numerical effect. As the rank increases, higher-order Jacobian directions associated with smaller singular values are included, making the reduced least-squares system severely ill-conditioned and amplifying finite-precision errors. This interpretation is supported by the double-precision results in Supplementary Fig. 7, where the high-rank loss increase observed in single precision largely disappears; however, double-precision computation incurs additional memory usage and computational cost. Consequently, enlarging the subspace introduces an inherent trade-off between increased corrective degrees of freedom and numerical stability, suggesting an optimal finite rank for LSR. Importantly, this instability is directly manifested in the loss behavior itself, providing a practical criterion for rank selection as the largest dimension for which the loss decreases monotonically. In addition to numerical conditioning, the memory and time required to form $\mathbf{AJV}$ and solve the rank-dependent reduced least-squares system provide practical constraints on the maximum rank used in LSR. Further details are provided in Supplementary Section 2.2.

These results show that LSR provides systematic post-training gains for trained neural operators, with the rank controlling the usable portion of the trained-state response space. The comparison with randomly initialized and trained Jacobians further confirms that nonlinear training shapes the local response space used by LSR.

**Physics-informed operator fine-tuning**

Unlike data-driven operator learning, physics-informed operator learning does not rely on explicit labeled data. Instead, supervision is provided directly through the governing equations together with initial and boundary conditions. As a result, the residual vector is naturally defined at a given prediction instance, rather than over an entire dataset. This structural distinction fundamentally alters how LSR is applied: in physics-informed settings, LSR can be performed directly on a single predicted solution by constructing the corresponding physical residual and solving a locally linearized correction problem. In this sense, LSR naturally operates as an instance-wise post-training refinement.

We evaluate this instance-wise LSR strategy on a collection of widely used benchmark physics-informed operator learning problems introduced by Wang et al.[32]. The benchmarks span a range of representative governing equations, including linear ordinary differential equations, reaction–diffusion systems, the Burgers equation, and advection equations, with uncertainty arising from random forcing terms, source terms, initial conditions, or spatially varying coefficients. In all cases, we adopt the original network architectures and training procedures to obtain baseline models, and then apply a single LSR step to the trained networks without any additional training. Quantitative results are summarized in Table 1. Across all benchmarks, LSR consistently reduces the physics residual loss at a small additional computational cost and yields stable improvements in test error, with error reductions ranging from approximately one to two orders of magnitude. These results demonstrate that, even in fully label-free and physics-constrained operator learning problems, LSR serves as an effective post-training refinement framework that exposes accuracy unattained by standard training.

**Table 1. LSR results for physics-informed operator fine-tuning.** All reported results are averaged over 300 instances. Loss denotes the composite physics residual loss used during training, and test error is evaluated against reference solutions.

| Governing law | Equation form | Random input | Baseline Loss | Baseline test error | LSR loss | LSR test error | Error reduction factor | Wall time (s) |
|---|---|---|---|---|---|---|---|---|
| Linear ODE | $\frac{dq}{dx} = u(x)$ | Forcing terms | $3.7\times10^{-5} \pm 4.2\times10^{-5}$ | $3.7\times10^{-3} \pm 3.5\times10^{-3}$ | $2.3\times10^{-10} \pm 2.0\times10^{-10}$ | $1.5\times10^{-5} \pm 1.3\times10^{-5}$ | ~240 | 0.04 |
| Diffusion reaction | $\frac{\partial q}{\partial t} = D\frac{\partial^2 q}{\partial x^2} + kq^2 + u(x)$ | Source terms | $6.9\times10^{-5} \pm 5.6\times10^{-5}$ | $5.6\times10^{-3} \pm 2.6\times10^{-3}$ | $8.5\times10^{-7} \pm 8.3\times10^{-7}$ | $5.7\times10^{-4} \pm 2.8\times10^{-4}$ | ~10 | 0.37 |
| Burgers' | $\frac{\partial q}{\partial t} + q\frac{\partial q}{\partial x} - \nu\frac{\partial^2 q}{\partial x^2} = 0$ | Initial conditions | $6.7\times10^{-4} \pm 5.2\times10^{-3}$ | $1.2\times10^{-2} \pm 1.5\times10^{-2}$ | $8.6\times10^{-9} \pm 3.9\times10^{-9}$ | $3.9\times10^{-4} \pm 2.2\times10^{-3}$ | ~30 | 6.90 |
| Advection | $\frac{\partial q}{\partial t} + u(x)\frac{\partial q}{\partial x} = 0$ | Variable coefficients | $4.9\times10^{-3} \pm 4.4\times10^{-3}$ | $2.2\times10^{-2} \pm 6.7\times10^{-3}$ | $3.1\times10^{-5} \pm 2.6\times10^{-5}$ | $3.1\times10^{-3} \pm 9.0\times10^{-4}$ | ~8 | 6.11 |

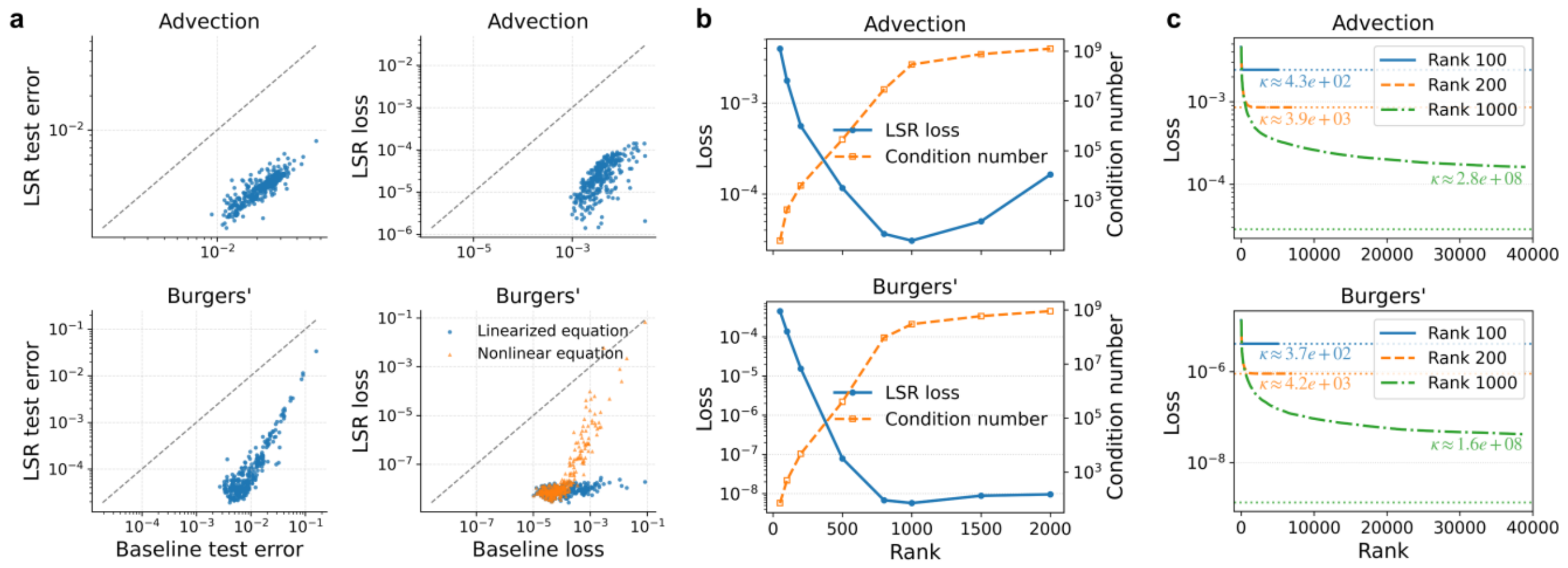

**Fig. 4. LSR behavior and conditioning effects in physics-informed operator fine-tuning. a,** Baseline and LSR prediction errors and losses for two representative equations evaluated across 300 prediction instances. **b,** Dependence of the LSR loss and the condition number of the reduced linearized least-squares system on the correction-subspace rank, averaged over 300 prediction instances. **c,** Convergence histories of full-batch L-BFGS applied to the reduced linearized least-squares problems at different ranks for representative instances. Horizontal reference lines indicate the loss levels achieved by the corresponding LSR solutions.

To further understand the behavior and limitations of LSR in physics-informed operator learning, we analyze its performance across different prediction instances and correction-subspace ranks, as shown in Fig. 4. In Fig. 4a, we compare baseline and LSR errors and losses for two representative equations, advection and Burgers. For the advection equation, LSR reliably reduces both error and loss across different instances. In contrast, for the Burgers equation, instances with larger baseline prediction errors tend to retain relatively large errors after LSR, exhibiting a strong correlation between baseline and LSR errors. Importantly, in physics-informed problems, LSR minimizes the linearized physical residual, and when the current prediction deviates substantially from the true solution of a nonlinear equation, the linearized residual can differ significantly from the original nonlinear residual. As a result, a substantial reduction in the linearized residual does not necessarily translate into a commensurate reduction in prediction error, a discrepancy that is clearly reflected in the loss comparison in Fig. 4a.

Figure 4b reveals a clear trade-off between subspace expressiveness and numerical stability. As the dimension of the linearized subspace (rank) increases, the LSR loss initially decreases rapidly, but rises again beyond an optimal range. In contrast, the condition number of the associated reduced least-squares system grows monotonically and sharply with rank. Consistent with this trend, Supplementary Fig. 10 shows that the norm of the subspace solution vector increases rapidly with rank and closely follows the growth of the condition number, indicating heightened sensitivity to numerical perturbations. These results confirm that the observed loss degradation at large ranks is driven by severe ill-conditioning of the reduced linear system.

Finally, Fig. 4c provides direct evidence for the origin of the accuracy plateaus observed in gradient-based optimization. For a representative prediction instance, we solve the reduced linearized least-squares problems obtained at the same trained network state using full-batch L-BFGS at different ranks, and compare the resulting convergence histories with the accuracy achieved by direct LSR solves, shown as horizontal reference lines. When the rank is small and the system remains moderately conditioned, L-BFGS converges to essentially the same accuracy level as the corresponding direct solution. However, as the rank increases and the condition number deteriorates, L-BFGS exhibits pronounced stagnation at substantially higher error levels and fails to approach the accuracy attained by the direct solve within a practical number of iterations. These results show that the same conditioning barrier observed in supervised regression also appears in physics-informed residual refinement: high-rank linearized subproblems remain convex, but their ill-conditioning limits what iterative solvers can practically attain.

**Rank-controlled LSR for noisy inverse problems**

Previous results show that LSR can expose additional attainable accuracy around trained neural networks. We next examine a complementary role of the LSR rank: capacity control in noisy inverse problems. Unlike noise-free refinement, inverse problems with noisy observations require a balance between reducing the data/physics residuals and avoiding noise amplification. This setting therefore provides a natural testbed for assessing whether the rank of the Jacobian-induced correction subspace can serve as an explicit and interpretable capacity-control parameter.

We use the heat inverse problem HInv from the PINNacle benchmark[33] as the test case. The task is to infer the spatially varying thermal conductivity $a(x,y)$ from noisy observations of the temperature field $u(x,y,t)$, while satisfying the heat equation $u_t - \nabla\cdot(a\nabla u) = f_s(x,y,t)$. The neural network simultaneously represents $u(x,y,t)$ and $a(x,y)$. We follow the standard PINNacle problem setting and use PINN-LRA[18] as the baseline training strategy, since it is reported as the best-performing method for this benchmark case. To study rank selection under noise, we extend the original benchmark by considering multiple noise levels and by splitting the noisy observations into 80% training data and 20% validation data. For each noise level and random seed, the validation set is used to select the baseline checkpoint and, after applying LSR, the subspace rank. The LSR residual uses the same loss components and final weights as the converged baseline PINN, and ranks from 0 to 3000 are scanned, where rank 0 denotes the original baseline without LSR correction. Importantly, rank selection uses only the noisy validation temperature data, without access to the clean temperature field or the ground-truth conductivity, mimicking a realistic noisy inverse-problem setting.

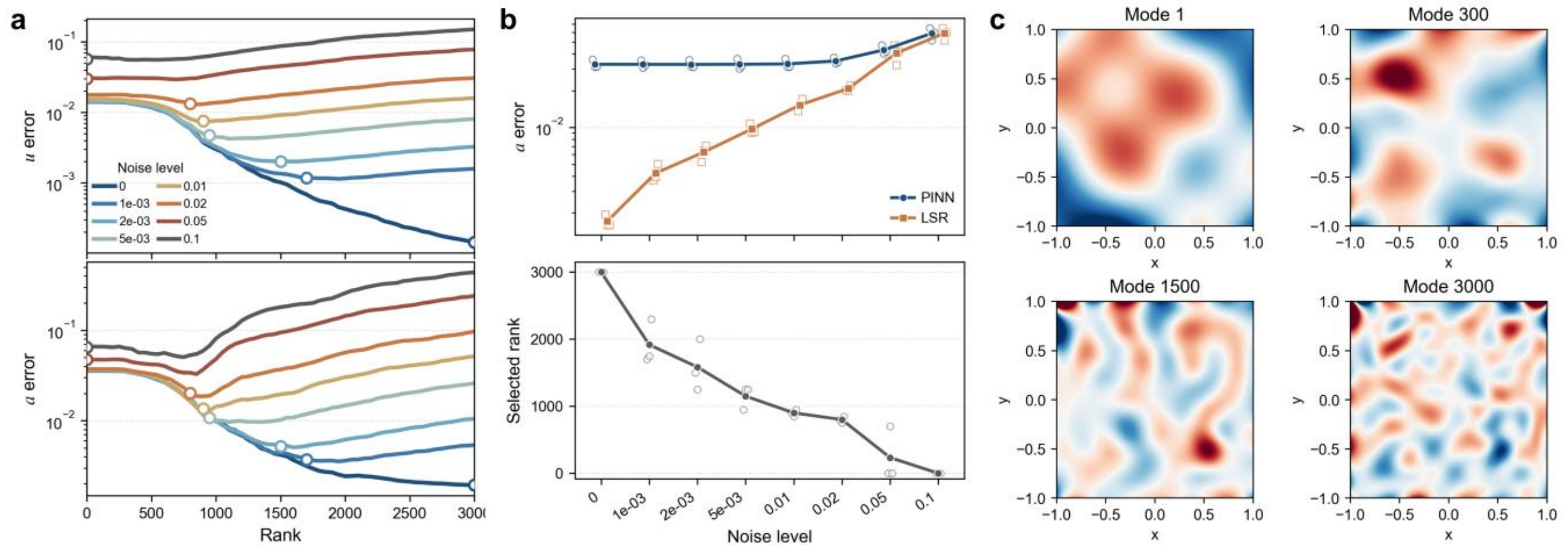

**Fig. 5. Rank-controlled LSR for noisy heat-conduction inverse problems. a,** Rank-dependent errors of the temperature field $u$ and thermal conductivity $a$ under different observation noise levels for one random seed. Circle markers indicate the ranks selected by the validation set. **b,** Comparison between the baseline PINN and validation-selected LSR across noise levels. The upper panel reports the conductivity error, and the lower panel shows the selected rank. **c,** Representative response modes of the temperature field induced by individual LSR subspace directions. For visualization, each mode is mean-subtracted and independently normalized to highlight its spatial structure and characteristic frequency content.

Figure 5a shows how the clean temperature error and conductivity error vary with the LSR rank under different noise levels. In the noise-free case, both errors decrease continuously as the rank increases, indicating that LSR can expose additional attainable accuracy around the trained PINN. Under noisy observations, however, the behavior becomes non-monotonic. Increasing the rank initially improves the reconstruction by adding effective correction directions, but overly large ranks can amplify noise and lead to larger errors in both $u$ and $a$, even when the LSR loss continues to decrease, as shown in Supplementary Fig. 12. The circle markers denote the ranks selected using only the noisy validation temperature data, not the clean temperature field or the ground-truth conductivity. Thus, the selected rank is directly tied to the observable state variable, whereas its effect on the inferred conductivity is assessed only a posteriori. This demonstrates that rank selection provides a practical, non-oracle capacity-control mechanism for noisy inverse problems.

Figure 5b summarizes the validation-selected LSR results and compares them with the baseline PINNs. Although the baseline error decreases as the observation noise is reduced, the low-noise results and the convergence histories in Supplementary Fig. 11 both show a clear saturation, indicating that the limiting factor is no longer only data quality but also the accuracy attainable by the trained PINN under standard optimization. In contrast, LSR further exploits the trained-state Jacobian-induced response space and achieves substantially lower errors under clean or low-to-moderate noise levels. For example, in the noise-free case, the mean relative $L_2$ error of $a$ is reduced from $3.29\times10^{-2}$ to $1.71\times10^{-3}$. As the noise level increases, the improvement becomes weaker and the validation-selected rank rapidly decreases, eventually reaching rank 0 at the highest noise level. This indicates that, under strong noise, the dominant limitation shifts from the optimization and approximation error of the trained PINN to the identifiability constraint imposed by noisy observations.

To further illustrate how the rank controls the complexity of the correction space, Fig. 5c visualizes representative response modes of the temperature field $u$. The low-index modes mainly exhibit large-scale, smooth structures, whereas higher-index modes become increasingly localized and oscillatory. This modal structure explains the capacity-control behavior observed in the rank scan: low-rank truncation retains stable large-scale corrections, while high-rank directions can represent finer structures but are also more likely to absorb observation noise. Therefore, the LSR rank can be interpreted as a data-driven complexity filter for noisy inverse problems.

# Discussion

## Accessing attainable accuracy beyond gradient-based training

This work reframes accuracy improvement in neural networks from a perspective complementary to prolonged gradient-based training. Rather than further updating parameters, we study the accuracy that remains accessible at a fixed trained network state through corrections confined to a low-dimensional Jacobian-induced subspace. This formulation also allows the attainable accuracy of training to be examined within a convex least-squares setting by contrasting the accuracy reached by iterative gradient-based optimization with that obtained via direct solvers on the same linearized residual problem. Although the accuracy attained by the linearized predictor should not be interpreted as a guaranteed attainable parameter update for the original nonlinear network, the discrepancy between iterative and direct solves in the local linearized problem provides evidence that standard gradient-based training may leave locally accessible correction capacity underexploited near the trained state.

LSR provides a practical mechanism for accessing this subspace-attainable accuracy. By constructing a compact correction basis and solving the reduced linearized residual problem directly, LSR yields a subspace-optimal local predictor without altering the original architecture, loss function, or training procedure. Across function approximation, data-driven operator learning, and physics-informed operator fine-tuning, the experiments show that trained-state linearized predictors can achieve substantially lower errors than the corresponding gradient-trained models. Thus, LSR acts as a post-training refinement tool that complements existing training methods by exposing accuracy that remains hidden at practical training plateaus.

The noisy inverse-problem results further show that the LSR rank is not only a computational truncation parameter, but also a control knob for the effective complexity of the correction space. In clean or low-noise regimes, larger ranks can reveal additional locally attainable accuracy. Under noisy observations, however, high-rank directions may absorb noise, whereas validation-selected low-rank corrections provide an explicit regularization mechanism. In this sense, rank-controlled LSR links accuracy enhancement with robustness by balancing correction capacity, numerical stability, and noise sensitivity.

**Conditioning-limited accuracy and the role of direct solvers**

Beyond the existence of locally attainable accuracy, our results indicate that numerical conditioning is a central factor limiting the accuracy reached by gradient-based training. By fixing the trained state and studying the associated linearized residual problem, LSR separates this effect from the nonconvexity of the original neural-network optimization. Even in the resulting convex least-squares problems, standard iterative optimizers can remain far above the direct-solve solution within practical computational budgets. The gap becomes more pronounced as the correction rank increases and the reduced system becomes increasingly ill-conditioned, revealing a conditioning barrier analogous to those encountered in classical numerical linear algebra.

Several additional diagnostics further support this interpretation. The rank-dependent linearized least-squares tests in Supplementary Section 2.1 show that, as the correction rank increases, the reduced system becomes increasingly ill-conditioned and the accuracy gap between L-BFGS and the direct LSR solution widens within the tested computational budget. Double precision alleviates part of the high-rank deterioration observed in direct LSR solves, indicating the role of finite-precision error amplification. By contrast, the L-BFGS results change little between single and double precision, suggesting that its plateau is mainly associated with the difficulty of iterative optimization in ill-conditioned reduced systems, rather than with floating-point precision alone. The finite-width empirical NTK comparison in Supplementary Section 3 further shows that full empirical NTK regression, despite using the entire tangent space, can be less stable than low-rank LSR because the full kernel system inherits and amplifies the spectral imbalance of the Jacobian.

**Scope and limitations**

Despite its effectiveness, the current LSR framework has an intrinsic scalability limitation. Its main cost lies not only in constructing the correction subspace, but also in assembling and solving the reduced linearized residual problem. Although the correction subspace can often be constructed with batched Jacobian actions and does not necessarily require the full training dataset, assembling the reduced least-squares system may still require residual Jacobian-vector products over the full residual set when the residual is defined over an entire training dataset. In this case, forming the reduced residual Jacobian, such as $\mathbf{AJV}$, requires a number of full-residual Jacobian-vector products proportional to the selected rank, followed by the assembly and direct solution of the reduced system. This cost grows with the dataset size, residual dimension, network size, and rank, making LSR unsuitable as a universal low-cost fine-tuning method for arbitrary large-scale models. Its present strength instead lies in medium-scale, accuracy-critical scientific machine learning problems, such as surrogate modeling of low-dimensional quantities of interest, data-driven and physics-informed operator refinement, and inverse problems, where the cost of a post-training high-accuracy correction is acceptable and reductions in prediction error, physical residual, or inverse-estimation error have direct value.

# Methods

## Linear residual model

Across supervised regression, data-driven operator learning, and physics-constrained PDE solving, training objectives can typically be expressed in terms of minimizing the norm of a residual vector,

$$\min_{\boldsymbol{\theta}} \frac{1}{2} \left\| \boldsymbol{f}(\boldsymbol{\theta}) \right\|_2^2, \tag{4}$$

where $\boldsymbol{f}(\boldsymbol{\theta}) \in \mathbb{R}^n$ denotes a residual vector constructed from model predictions evaluated at data samples, observations, or physical constraints. In this work, we focus on a trained parameter state $\boldsymbol{\theta}_0$ and construct a Jacobian-induced linear residual model

$$\boldsymbol{f}(\boldsymbol{\theta}_0) + \mathbf{G}\Delta\boldsymbol{\theta}, \mathbf{G} \equiv \left. \frac{\partial \boldsymbol{f}}{\partial \boldsymbol{\theta}} \right|_{\theta_0}, \tag{5}$$

with the objective of identifying $\Delta\boldsymbol{\theta}$ that reduces the residual norm within this linearized model.

For notational consistency across learning paradigms, we formally write the residual Jacobian as $\mathbf{G} = \mathbf{AJ}$. In standard supervised learning, $\mathbf{A} = \mathbf{I}$. In physics-informed learning based on automatic differentiation, however, both $\boldsymbol{f}$ and $\boldsymbol{q}$ depend directly on $\boldsymbol{\theta}$, and $\mathbf{A}$ and $\mathbf{J}$ need not be available separately. In practice, only the action of the relevant Jacobian operators is required: the action of $\mathbf{J}$ for constructing the parameter subspace, and the action of the combined operator $\mathbf{G} = \mathbf{AJ}$ for evaluating residual corrections. We therefore retain the notation $\mathbf{AJ}$ without implying that either factor must be explicitly constructed.

## Subspace reduction and LSR

Directly solving the linear residual system in the full parameter space is generally infeasible due to the extremely high dimensionality of $\boldsymbol{\theta}$. We therefore restrict the parameter correction to a low-dimensional subspace,

$$\Delta\boldsymbol{\theta} = \mathbf{V}\boldsymbol{y}, \mathbf{V} \in \mathbb{R}^{m\times r}, r \ll m \tag{6}$$

where the columns of $\mathbf{V}$ form a basis of the correction subspace, capturing parameter directions that most strongly influence variations of the network output at $\boldsymbol{\theta}_0$.

The subspace basis $\mathbf{V}$ is constructed using RSVD. Specifically, a Gaussian random matrix $\boldsymbol{\Omega}$ is drawn and the product $\mathbf{J\Omega}$ is computed via Jacobian–vector products. A QR factorization yields an approximate orthonormal basis for the output space. We then form the projected Jacobian operator $\mathbf{B} = \mathbf{Q}^T\mathbf{J}$ and compute its small-scale SVD. The leading right singular vectors of $\mathbf{B}$ approximate the dominant right singular directions of $\mathbf{J}$, forming the columns of $\mathbf{V}$.

Restricting the linear residual model to the subspace leads to the reduced linear system

$$\mathbf{AJV}\boldsymbol{y} = -\boldsymbol{f}(\boldsymbol{\theta}_0) \tag{7}$$

of size $n \times r$. This system is solved using a direct linear least-squares solver (e.g. QR decomposition), yielding the LSR solution

$$\Delta\boldsymbol{\theta} = \mathbf{V}\boldsymbol{y}. \tag{8}$$

The entire procedure involves no iterative optimization and does not require explicit formation or storage of $\mathbf{J}$ or $\mathbf{G}$. Instead, it relies on Jacobian–vector products to construct the correction subspace and to assemble the reduced residual Jacobian $\mathbf{AJV}$, followed by a small-scale direct least-squares solve. A detailed summary of the algorithm is provided in Algorithm 1.

**Algorithm 1: Linearized Subspace Refinement (LSR)**

**Input:** Trained parameters $\boldsymbol{\theta}_0$, residual function $\boldsymbol{f}(\boldsymbol{\theta})$, rank *r*, oversampling parameter *p*, access to Jacobian–vector products $\mathbf{J}\boldsymbol{v}$, vector–Jacobian products $\mathbf{J}^T\boldsymbol{v}$; residual Jacobian–vector products $(\mathbf{AJ})\boldsymbol{v}$.

1. Evaluate residual vector $\boldsymbol{f}_0 = \boldsymbol{f}(\boldsymbol{\theta}_0)$.
2. Draw a Gaussian random matrix $\boldsymbol{\Omega} \in \mathbb{R}^{m\times(r+p)}$, where *m* is the dimension of parameter vector $\boldsymbol{\theta}_0$
3. Compute $\mathbf{Y} = \mathbf{J\Omega}$ using Jacobian–vector products.
4. Compute a QR factorization $\mathbf{Y} = \mathbf{QR}$
5. Form the projected Jacobian operator $\mathbf{B} = \mathbf{Q}^T\mathbf{J}$ and compute its SVD $\mathbf{B} = \tilde{\mathbf{U}}\tilde{\boldsymbol{\Sigma}}\tilde{\mathbf{V}}^T$
6. Construct the subspace basis $\mathbf{V} \leftarrow \tilde{\mathbf{V}}(:,1:r)$ (optionally scaled by $\tilde{\boldsymbol{\Sigma}}_r^{-1}$ for preconditioning).
7. Assemble the reduced linearized residual system $\mathbf{AJV}\boldsymbol{y} = -\boldsymbol{f}_0$.
8. Solve the reduced system using QR decomposition to obtain $\boldsymbol{y}^*$.
9. Set $\Delta\boldsymbol{\theta}^* = \mathbf{V}\boldsymbol{y}^*$

**Output:** LSR prediction $\boldsymbol{q}_{LSR} = \boldsymbol{q}(\boldsymbol{\theta}_0) + \mathbf{J}\Delta\boldsymbol{\theta}^*$, and LSR residual $\boldsymbol{f}_{LSR} = \boldsymbol{f}(\boldsymbol{\theta}_0) + \mathbf{AJ}\Delta\boldsymbol{\theta}^*$

**Stationarity and zero-correction behavior**

If $\boldsymbol{\theta}_0$ is a first-order stationary point of the full-training-set residual minimization problem, then the linearized residual model at $\boldsymbol{\theta}_0$ admits no first-order descent direction, and LSR yields a zero correction. This outcome is expected: at stationarity, first-order information cannot identify a locally improving direction for the linearized least-squares objective. In practical training, however, iterative gradient-based optimization may terminate before reaching exact stationarity under finite iteration and tolerance budgets, even when the locally linearized subproblem is convex. Consequently, the correction subspace can retain nonzero reducible components, so LSR may still produce a nontrivial refinement of the linearized predictor. Additional discussion of this boundary case and its implications for linearized correction spaces is provided in Supplementary Section 5.

**Batch LSR for large-scale datasets**

In data-driven operator learning, residuals are often defined over entire training datasets, leading to extremely large residual dimension $n$. In such settings, intermediate quantities such as $\mathbf{J\Omega}$ and $\mathbf{AJV}$ cannot be formed or stored explicitly at the dataset level. We therefore employ a batch LSR strategy, in which residuals and Jacobian actions are computed over data batches and accumulated in a streaming fashion to assemble the reduced linear system. This approach avoids explicit storage of full dataset–level matrices while preserving the correctness of the reduced linear solve. Implementation details and numerical considerations are provided in Supplementary Section 1.

## Data availability

The data used in this study were either obtained from previously published works or generated using the numerical procedures described in the manuscript. Data supporting the findings of this study are available within the Article and its Supplementary Information. Additional data required to reproduce the reported results are provided together with the source code.

## Code availability

The source code used to generate the results reported in this study is publicly available at Zenodo (https://doi.org/10.5281/zenodo.20623071) and at the corresponding GitHub repository (https://github.com/Cao-WenBo/LinearizedSubspaceRefinement).

## Acknowledgements

This work was supported by the National Natural Science Foundation of China (No. 92152301).

## Author contributions

W.C. and W.Z. conceived the study and developed the methodology. W.C. developed the software and performed the investigation and visualization. W.Z. supervised the study and acquired funding. W.C. wrote the original draft. W.C. and W.Z. reviewed and edited the manuscript.

## Competing interests

The authors declare no competing interests.

# Linearized subspace refinement framework to expose hidden accuracy in trained neural networks

## Supplementary Information

## 1. Additional methodological details

### 1.1 Verification of randomized SVD for constructing the LSR correction subspace

The LSR subspace is constructed to approximate the dominant parameter-space directions induced by the network Jacobian. Since the Jacobian $\mathbf{J}$ is generally rectangular, these directions correspond to the leading right singular vectors of $\mathbf{J}$. In this work, randomized singular value decomposition (RSVD) is used to approximate this dominant right singular subspace in a matrix-free manner through Jacobian-vector and vector-Jacobian products, without explicitly forming or storing the full Jacobian. The Gaussian random matrix $\mathbf{\Omega}$ is used as a sketching matrix to probe the dominant Jacobian action and approximate the leading right singular subspace.

To verify the accuracy and efficiency of this RSVD-based construction, we conduct a controlled comparison using a reduced-scale version of the function approximation problem in Figure 2 of the main text. The target function and training procedure are kept unchanged, while the network and dataset sizes are reduced so that the full Jacobian can be explicitly constructed. Specifically, the network contains four hidden layers with 32 neurons per layer, and the dataset contains 2000 training points and 1000 test points. This setting makes it possible to compute the exact SVD of the full Jacobian and use it as a reference. Starting from the same trained base model, we compare three subspace construction methods. The first method is exact SVD, in which the full Jacobian J is explicitly formed and its leading right singular vectors are used as the subspace basis. The second method is a matrix-free iterative top-r approximation, implemented as block subspace iteration on $\mathbf{J}^T\mathbf{J}$. This method avoids explicit Jacobian formation and approximates the dominant right singular subspace by repeatedly applying $\mathbf{J}^T\mathbf{J}$ through JVP/VJP operations; 20 subspace iterations are used in the present test. The third method is the RSVD method used in the main experiments, with an oversampling parameter $p = 50$.

Supplementary Fig. 1 compares the LSR test errors obtained using these three subspace bases. The three curves nearly overlap across the tested ranks, showing that RSVD achieves essentially the same refinement accuracy as exact SVD and matrix-free iterative top-$r$ approximation. The corresponding response modes are also highly consistent, as shown in Supplementary Fig. 2, indicating that the three methods identify nearly the same dominant correction subspace and produce similar Jacobian-induced response modes. Only minor differences are observed in some higher-order modes, where the associated singular directions are less dominant and more sensitive to numerical approximation. The main difference lies in computational cost. At rank $r = 400$, exact SVD requires approximately 16.37 s, the matrix-free iterative top-$r$ method requires approximately 1.22 s, whereas RSVD requires only approximately 0.18 s.

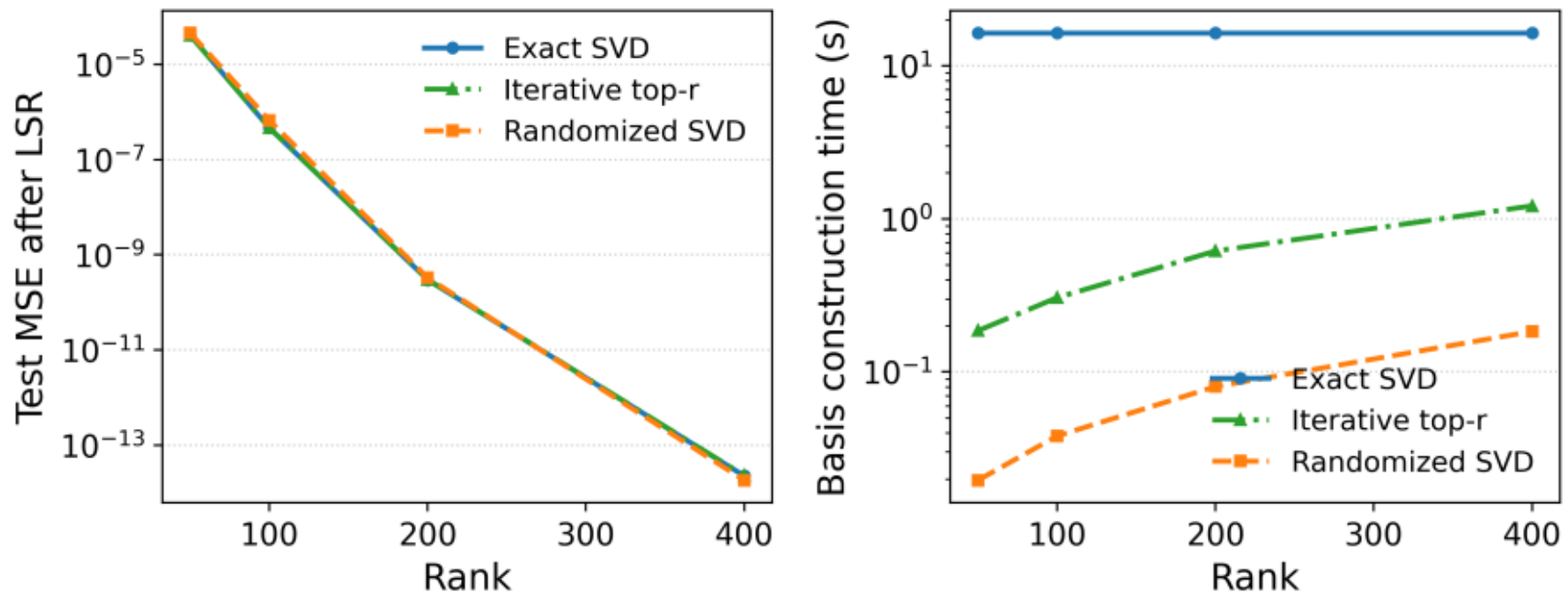

Supplementary Fig. 1. Accuracy and construction cost of different subspace construction methods.

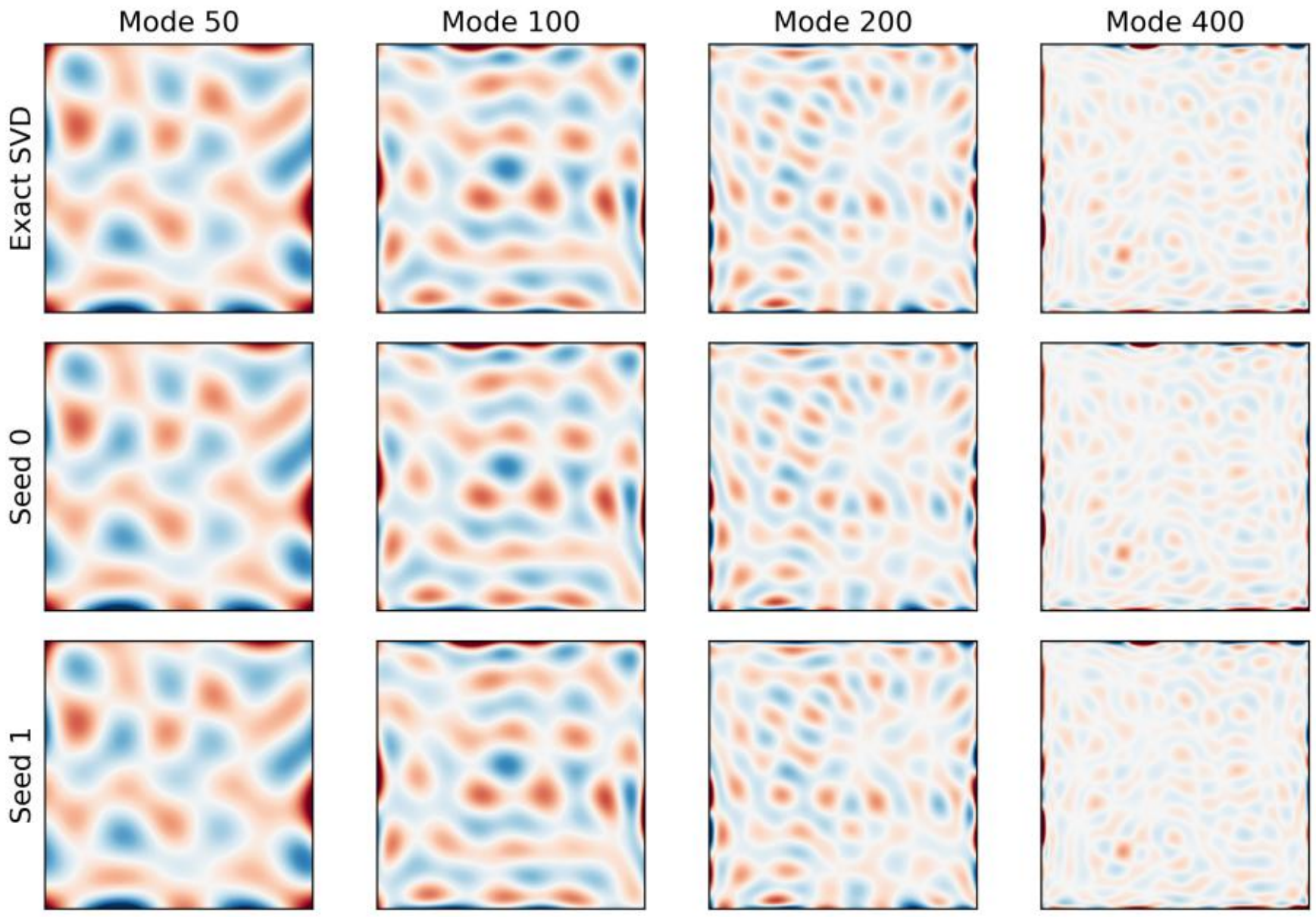

Supplementary Fig. 2. Comparison of response modes obtained by exact SVD and randomized SVD.

Exact SVD is useful as a reference but requires explicit Jacobian formation and dense matrix decomposition, which becomes impractical for large residual dimensions or large neural networks. Matrix-free iterative top-$r$ methods avoid storing $\mathbf{J}$, but they rely on sequential subspace iterations. Their cost therefore grows with the number of iterations or convergence tolerance, and their parallel efficiency is limited by this sequential dependence. In contrast, RSVD uses a fixed sketch size and batched Jacobian actions. The dominant correction subspace can therefore be approximated through highly parallel JVP/VJP computations, making RSVD substantially more efficient and scalable while preserving the refinement accuracy.

### 1.2 Choice of subspace basis: output space versus residual space

In the standard LSR formulation presented in the main text, the subspace basis $\mathbf{V}$ is constructed from the Jacobian of the network output $q(\boldsymbol{\theta})$, that is, by approximating the leading right singular subspace of $\mathbf{J} = \partial \boldsymbol{q} / \partial \boldsymbol{\theta}$ using RSVD. For operator-constrained problems, where the residual takes the form $\boldsymbol{f}(\boldsymbol{\theta})$, one may alternatively construct the subspace from the residual Jacobian $\mathbf{G} = \partial \boldsymbol{f} / \partial \boldsymbol{\theta} = \mathbf{AJ}$, thereby targeting the residual space directly. In practice, however, we did not observe systematic or practically significant differences in refinement quality between output-space and residual-space constructions. By contrast, constructing the residual-space basis typically incurs a substantially higher computational cost, as it requires repeated Jacobian-vector products involving the composite operator $\mathbf{AJ}$ during subspace construction. For this reason, we recommend constructing $\mathbf{V}$ from the output Jacobian $\mathbf{J}$ in most practical settings.

### 1.3 Subspace preconditioning and its effect on solvers

In LSR, the subspace basis $\mathbf{V}$ is optionally preconditioned using the inverse singular values obtained from the reduced SVD, that is, replacing $\mathbf{V}$ by $\mathbf{V}\boldsymbol{\Sigma}^{-1}$. This preconditioning improves the conditioning of the reduced subspace and is equivalent to whitening the dominant Jacobian directions.

Empirically, we observe that this preconditioning has little effect on the accuracy or robustness of direct solvers used in LSR. However, it can significantly accelerate convergence when gradient-based optimization methods are applied within the reduced subspace. In the main text, where the objective is to isolate and analyze the effect of subspace conditioning on the attainable accuracy of gradient-based training, we therefore deliberately omit this preconditioning when solving reduced linear systems at different ranks. This choice is also conceptually consistent. The preconditioning matrix arises naturally from the reduced-basis construction process, and when no subspace reduction is performed, such a preconditioner is not available.

### 1.4 Batch LSR for large-scale datasets

For data-driven operator learning, residuals are often defined over very large training datasets, making it infeasible to form Jacobian-related quantities over the full dataset at once. To address this challenge, we implement a **batch LSR strategy** that evaluates Jacobian actions and assembles the reduced linear system incrementally in a streaming fashion, as shown in Algorithm 1.

**Algorithm 1: Batch Linearized Subspace Refinement (batch-LSR)**

**Input:** Trained parameters $\boldsymbol{\theta}_0$, residual function $\boldsymbol{f}(\boldsymbol{\theta})$, rank *r*, oversampling parameter *p*, training data loader, access to Jacobian–vector products $\mathbf{J}\boldsymbol{v}$, vector–Jacobian products $\mathbf{J}^T\boldsymbol{v}$; residual Jacobian–vector products $(\mathbf{AJ})\boldsymbol{v}$.

**Step 1: Subspace construction**

1. Draw a Gaussian random matrix $\boldsymbol{\Omega} \in \mathbb{R}^{m\times(r+p)}$, where *m* is the dimension of parameter vector $\boldsymbol{\theta}_0$
2. Initialize an accumulator $\mathbf{H} = \mathbf{0} \in \mathbb{R}^{(r+p)\times m}$
3. For each data batch:

    Compute $\mathbf{Y} = \mathbf{J}\boldsymbol{\Omega}$ using Jacobian–vector products.

    Perform a reduced QR decomposition $\mathbf{Y} = \mathbf{QR}$. (If the final batch is smaller than required for QR decomposition due to fixed batch sizing, it is omitted)

Accumulate $\mathbf{H} \leftarrow \mathbf{H} + \mathbf{Q}^T \mathbf{J}$ .

4. Compute the compact SVD $\mathbf{H} = \tilde{\mathbf{U}}\tilde{\boldsymbol{\Sigma}}\tilde{\mathbf{V}}^T$, and construct the correction subspace $\mathbf{V} \leftarrow \tilde{\mathbf{V}}(:,1:r)$ or $\mathbf{V} \leftarrow \tilde{\mathbf{V}}(:,1:r)\tilde{\boldsymbol{\Sigma}}_r^{-1}$

**Step 2: Streaming assembly of the reduced linear system**

5. Initialize $\mathbf{R} = \mathbf{0} \in \mathbb{R}^{r \times r}$, $\boldsymbol{z} = \mathbf{0} \in \mathbb{R}^r$

6. For each data batch:

Compute $\mathbf{Y} = \mathbf{AJV}$ using residual Jacobian-vector products, and evaluate the batch residual vector $\tilde{\boldsymbol{b}} = -\boldsymbol{f}(\boldsymbol{\theta}_0)$

Form the stacked matrix and right-hand side

$$\mathbf{M} \leftarrow \begin{bmatrix} \mathbf{R} \\ \mathbf{Y} \end{bmatrix}, \boldsymbol{b} \leftarrow \begin{bmatrix} \boldsymbol{z} \\ \tilde{\boldsymbol{b}} \end{bmatrix}$$

Compute a reduced QR factorization $\mathbf{M} = \mathbf{Q}_m \mathbf{R}_{new}$

Update $\mathbf{R} \leftarrow \mathbf{R}_{new}, \boldsymbol{z} \leftarrow \mathbf{Q}_m^T \boldsymbol{b}$

**Step 3: Parameter correction**

7. Solve the reduced triangular system $\mathbf{R}\boldsymbol{y} = \boldsymbol{z}$ to obtain $\boldsymbol{y}^*$.

8. Set $\Delta\boldsymbol{\theta}^* = \mathbf{V}\boldsymbol{y}^*$

**Output:** LSR prediction $\boldsymbol{q}_{LSR} = \boldsymbol{q}(\boldsymbol{\theta}_0) + \mathbf{J}\Delta\boldsymbol{\theta}^*$, and LSR residual $\boldsymbol{f}_{LSR} = \boldsymbol{f}(\boldsymbol{\theta}_0) + \mathbf{AJ}\Delta\boldsymbol{\theta}^*$

Once the subspace basis **V** is fixed, the batch QR–based assembly and solution of the reduced least-squares problem is mathematically equivalent to the corresponding non-batch formulation using the full dataset, and yields the same reduced normal equations. The use of batch processing only affects the construction of the subspace basis **V**, as the Jacobian actions are accumulated incrementally across data batches. In practice, we find that this difference in subspace construction has no noticeable impact on the refinement quality or the attainable accuracy. In addition, although LSR relies on a randomized Gaussian random matrix $\boldsymbol{\Omega}$, we observe very low variance across different random realizations in practice, and therefore do not further elaborate on this aspect.

### 1.5 Practical considerations

To reduce memory and computational costs during subspace construction, it is not necessary to use the entire training dataset. It suffices that the residual dimension exceeds the target rank. Empirically, we find that using data batches whose combined residual dimension is approximately $r$+1000 is sufficient to obtain a stable and representative refinement subspace.

In batch LSR, the maximum matrix size—and hence the peak memory footprint—is primarily governed by the parameter dimension *m*. For very large neural networks, additional memory savings can be achieved by randomly subsampling network parameters during the construction of the subspace basis **V**.

These implementation details improve the practicality of LSR for larger operator-learning problems, while preserving the reduced linear solve used in the main experiments.

## 2. Training details and additional results

### 2.1 Function approximation

For the function approximation task, we use a fully connected neural network with 6 hidden layers of width 128 and tanh activation functions. The training and test datasets consist of 8,000 and 2,000 randomly sampled points, respectively. The baseline network parameters are optimized using the Adam optimizer, together with a ReduceLROnPlateau learning-rate scheduling strategy. The batch size is set to 32 throughout training. The gradient-based training curves for this task are reported in the main text. For all LSR experiments in this setting, the refinement subspace dimension is fixed to $r = 1000$.

The main text reports optimizer-dependent convergence for the linearized least-squares problem; here we add a hyperparameter sensitivity study. Supplementary Fig. 3 further shows the train MSE convergence for the convex least-squares problem obtained from local linearization, solved using Adam and L-BFGS optimizers with different hyperparameters. Although the linearized objective is convex and admits an accurate solution via a direct least-squares solve in the reduced subspace, iterative optimization methods exhibit clear accuracy plateaus.

For Adam, larger learning rates yield faster initial convergence, but all tested learning rates converge to nearly identical asymptotic error levels after $10^6$ iterations. These asymptotic errors remain orders of magnitude larger than those achieved by the direct solver. For L-BFGS, increasing the history size accelerates early convergence through improved curvature approximation, but does not improve the final attainable accuracy. All tested history sizes converge to similar error levels, again substantially above the direct-solve solution. Together, these results show that even for convex quadratic problems, iterative optimization methods can fail to reach the accuracy accessible to direct linear solvers, highlighting numerical ill-conditioning as a key bottleneck.

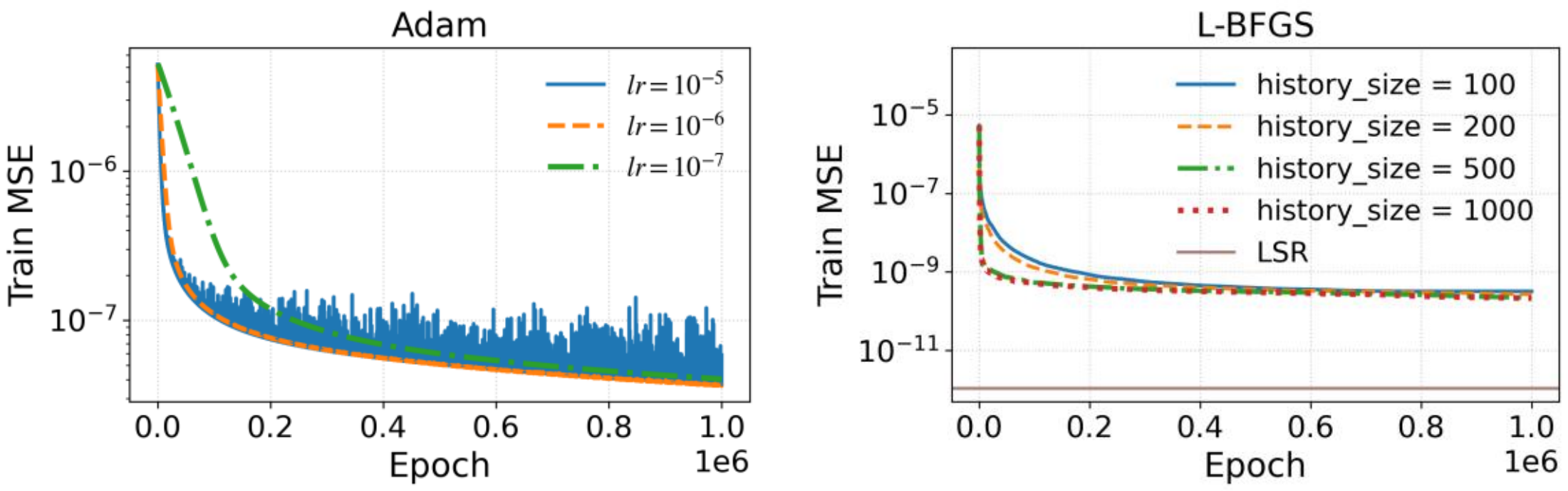

Supplementary Fig. 3. Train MSE convergence for a convex least-squares problem induced by local linearization.

To further examine the role of numerical ill-conditioning in limiting attainable accuracy, we solve the local linearized least-squares problems at different subspace rank. For each rank, we compare the LSR solution with the result obtained by applying L-BFGS to the same reduced convex least-squares problem for $10^6$ iterations. Both single-precision and double-precision computations are tested.

Supplementary Fig. 4 shows that, at low ranks, L-BFGS reaches nearly the same training MSE as the LSR solution. In this regime, the reduced system is only mildly ill-conditioned, and iterative optimization can effectively approach the direct-solve accuracy. As the rank increases, however, the condition number of the reduced system grows rapidly. Correspondingly, the gap between L-BFGS and the LSR solution becomes increasingly pronounced. This behavior demonstrates that, even after nonconvexity has been removed by local linearization, severe ill-conditioning can still prevent gradient-based iterative methods from reaching the accuracy attainable by a direct least-squares solve within a practical computation budget.

The comparison between single and double precision provides further insight. The L-BFGS results in single and double precision are nearly identical across ranks, indicating that simply increasing floating-point precision does not substantially improve the accuracy reached by iterative optimization within the given iteration budget. In contrast, the LSR solution benefits from double precision at high ranks, reaching lower training MSE than its single-precision counterpart. This contrast suggests that the observed optimization plateau is not merely a finite-precision artifact; rather, it reflects the difficulty of iterative methods in exploiting severely ill-conditioned correction directions. Direct solvers can access these directions more effectively, although their ultimate accuracy is still affected by numerical precision when the reduced system becomes highly ill-conditioned.

These results reinforce the central point that the bottleneck studied in this work is not only the expressive capacity of the local linearized model, but also the numerical attainability of high-accuracy solutions under finite precision and finite computational budgets.

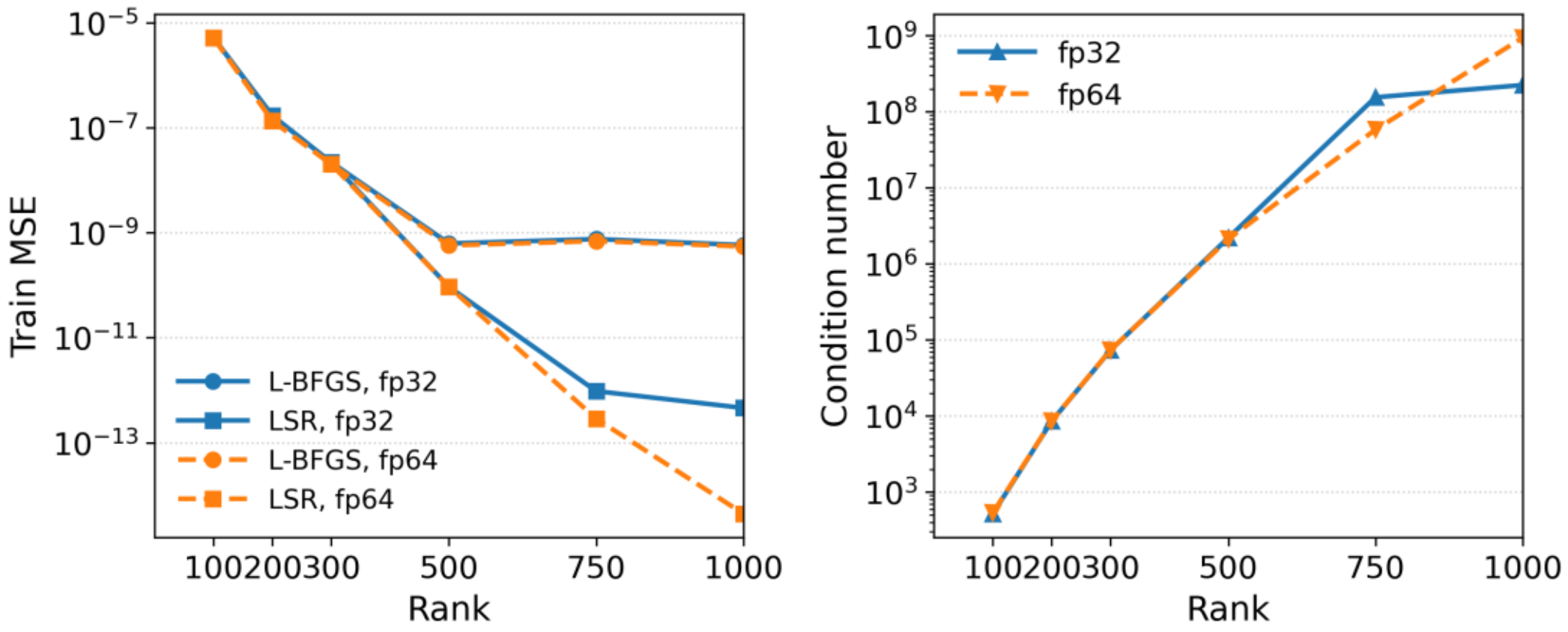

Supplementary Fig. 4. Rank-dependent conditioning effects in the linearized least-squares problem.

## 2.2 Data-driven operator learning

The data-driven operator learning experiments in this section follow the benchmark setup and implementation provided in Ref.[1], together with its publicly available code repository at https://github.com/yaohua32/Deep-Neural-Operators-for-PDEs/tree/main. Our modifications are limited to integrating LSR into the existing training and evaluation pipeline.

We consider the one-dimensional, time-dependent viscous Burgers' equation, a standard benchmark for evaluating neural operator methods, $q_t + qq_x = \nu q_{xx}, (x,t) \in [-1,1] \times [0,1]$, with initial condition $q(x,t=0) = a(x)$ and periodic boundary conditions. The viscosity coefficient is set to $\nu = 0.1/\pi$. In this setting, the PDE input corresponds to the initial condition, and the learning objective is to predict the final solution $q(x,t=1)$.

The initial conditions are generated by sampling from a Gaussian process $GP(0, 49^2(-\Delta + 49I)^{-2})$, evaluated at 128 uniformly spaced grid points on the interval $[-1,1]$. Consequently, both the input and output of the neural operator are 128-dimensional vectors. The training dataset consists of 1,000 input–output pairs, and an additional 50 samples are used for testing.

For DeepONet, both the branch and trunk networks are implemented as fully connected networks with four hidden layers of width 128. The MultiONet architecture and associated hyperparameters follow the original code; further details can be found in the corresponding reference and its open-source implementation. In addition to the commonly used ReLU and tanh activation functions, we also consider the Tanh–Sin activation function defined as $tanh\left(sin\left(\pi\left(x+1\right)\right)\right)+x$. Additional implementation details are provided in the cited literature and accompanying code.

Supplementary Fig. 5 further reports the loss values and test errors as functions of the subspace rank for all six test cases considered. To reduce memory consumption, only a randomly selected half of the network parameters are used during subspace construction in all experiments. Across different network architectures and activation functions, we consistently observe systematic accuracy improvements with LSR, indicating robustness to model design choices. At very large subspace dimension, however, numerical instability can arise due to amplified ill-conditioning in the reduced least-squares system, which may limit further gains.

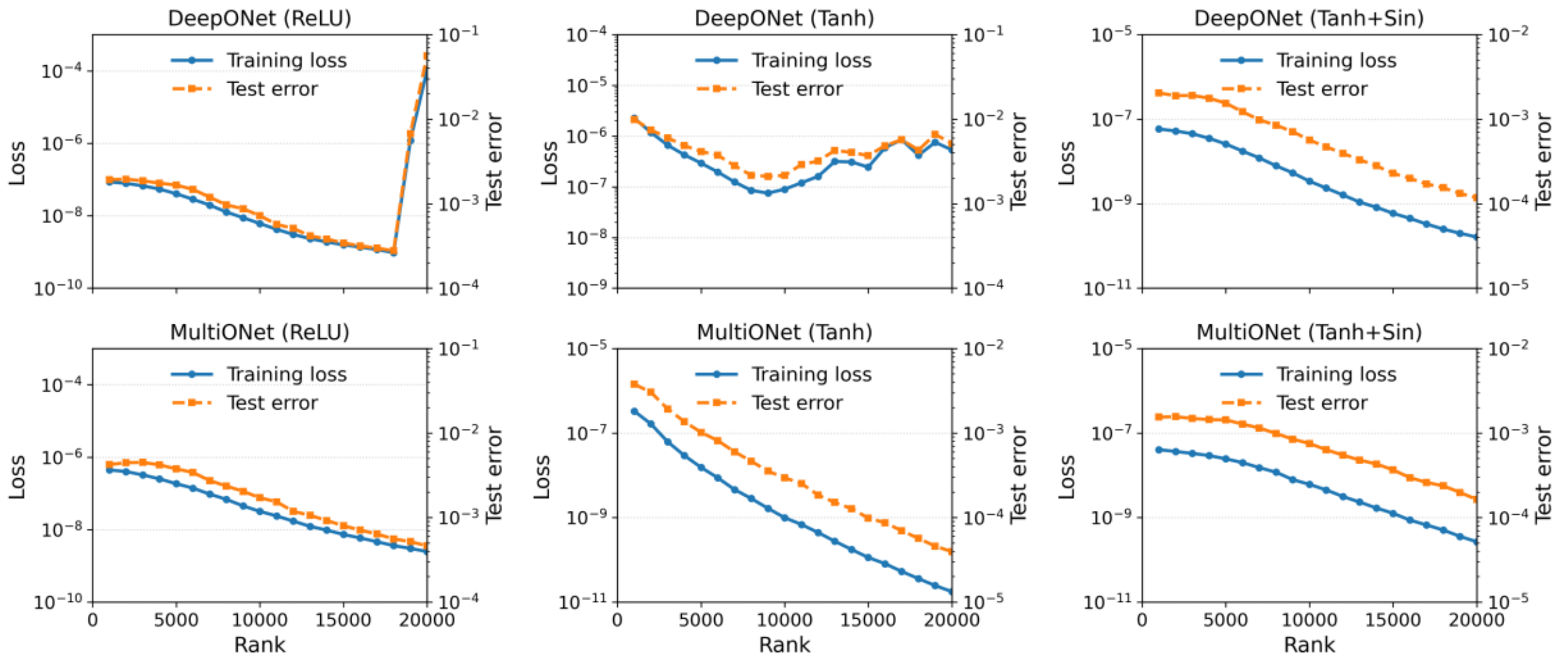

Supplementary Fig. 5. Training loss and test error versus subspace rank for data-driven operator learning.

To distinguish the effect of the trained base prediction from the effect of the Jacobian-induced response space, we perform a controlled comparison using the same trained base prediction and residual. Specifically, starting from the trained prediction $q(\boldsymbol{\theta}_{\text{train}})$, we construct two LSR corrections using different Jacobians: $q = q(\boldsymbol{\theta}_{\text{train}}) + \mathbf{J}(\boldsymbol{\theta}_{\text{init}})\Delta\boldsymbol{\theta}$ and $q = q(\boldsymbol{\theta}_{\text{train}}) + \mathbf{J}(\boldsymbol{\theta}_{\text{train}})\Delta\boldsymbol{\theta}$. The first case uses the Jacobian evaluated at random initialization to correct the trained residual, whereas the second uses the Jacobian evaluated at the trained state. Since the base prediction and residual are identical in the two cases, any difference in the LSR result reflects the effect of the Jacobian-induced response space rather than the base prediction itself.

Supplementary Fig. 6 shows that the correction based on the trained Jacobian consistently achieves lower loss and test error than the correction based on the randomly initialized Jacobian across the tested ranks. This indicates that training not only improves the base prediction, but also reshapes the local Jacobian-induced response space of the finite-width network, making the trained-state Jacobian more effective for correcting the current residual.

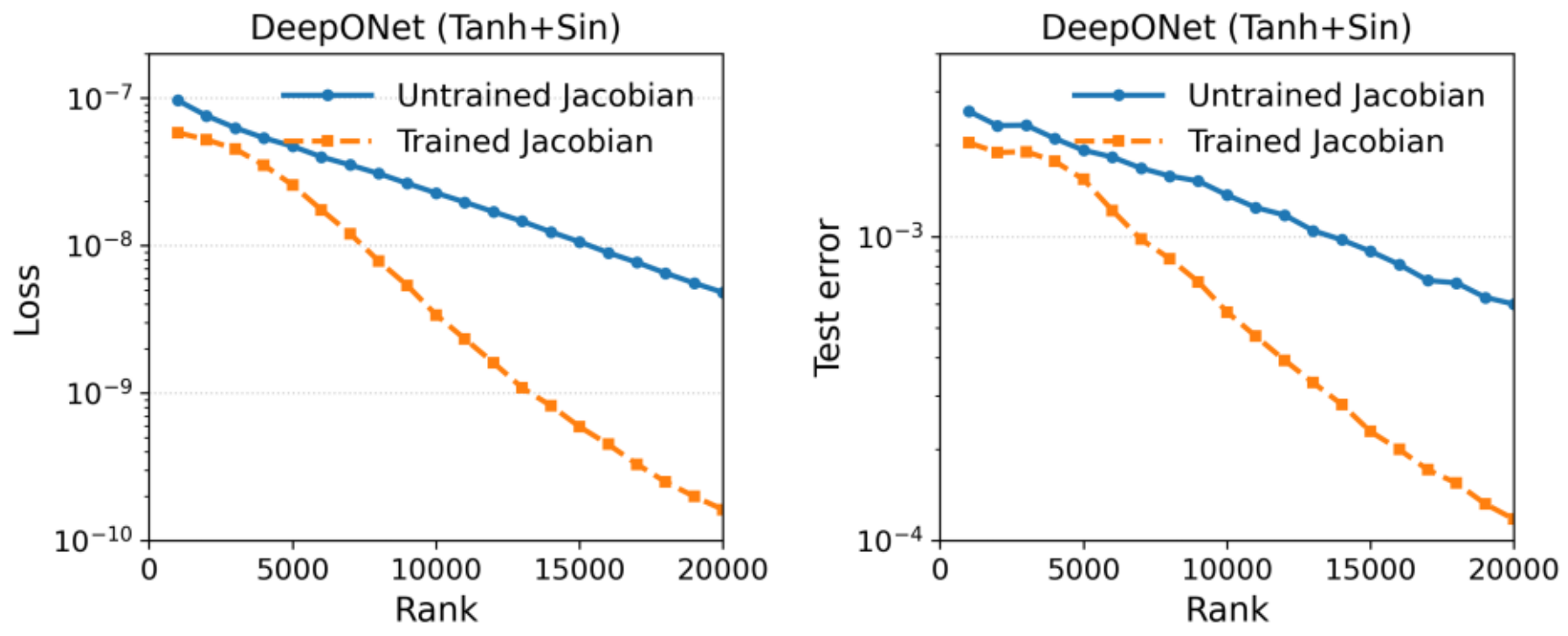

Supplementary Fig. 6. Controlled comparison between randomly initialized and trained Jacobians for correcting the same trained residual.

To further verify that the high-rank deterioration is mainly caused by numerical ill-conditioning, we repeated the rank-dependent LSR computation for the trained DeepONet model with the Tanh activation using double precision. As shown in Supplementary Fig. 7, the double-precision results continue to decrease over the same rank range, and the high-rank loss increase observed in single precision largely disappears. This comparison confirms that the deterioration at high ranks is primarily a finite-precision effect induced by severe ill-conditioning of the reduced least-squares system. Nevertheless, double-precision computation substantially increases memory usage and computational time, which made larger-rank calculations, such as rank 16000, infeasible in our setting. Therefore, in practical applications, the choice of rank should balance correction capacity, numerical stability, and computational cost.

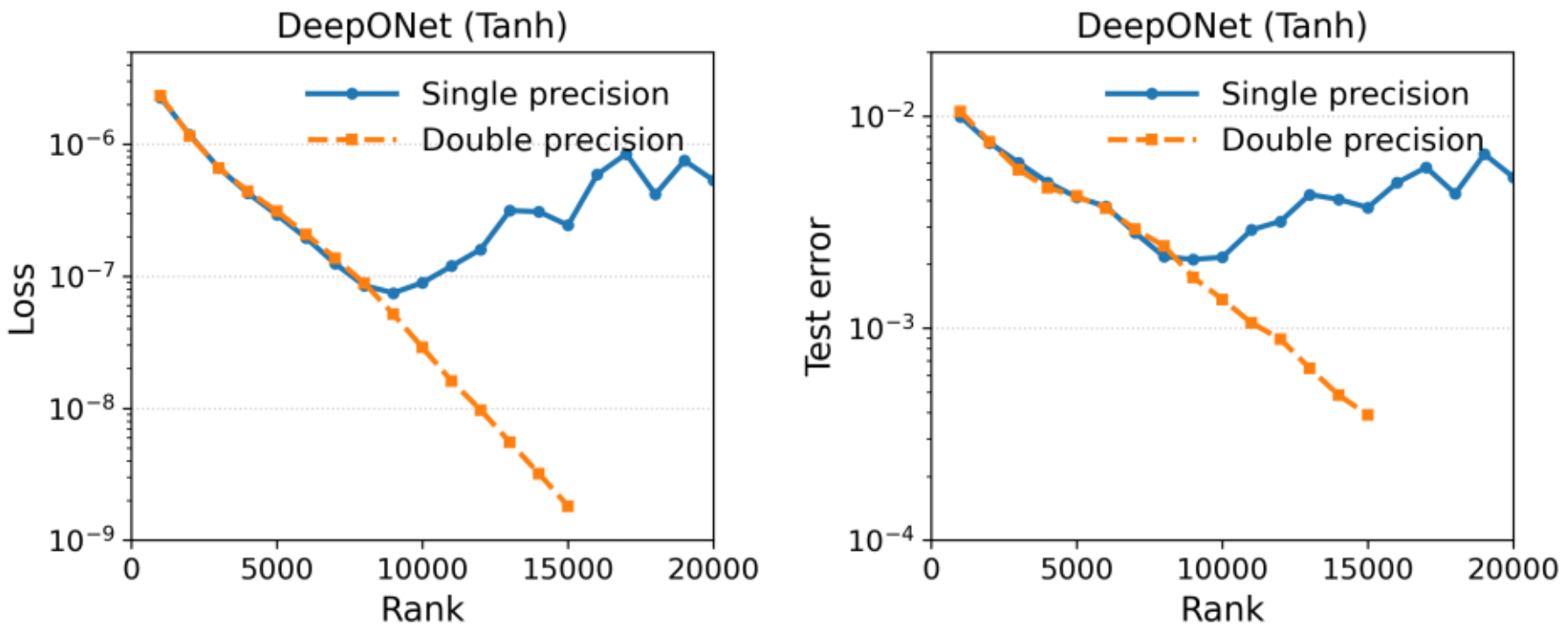

Supplementary Fig. 7. Effect of numerical precision on high-rank LSR deterioration.

Supplementary Fig. 8 reports the memory consumption and wall-clock time of single-precision LSR as functions of the subspace rank, highlighting the extent to which the method is constrained by computational resources. As expected, we observe an approximately linear growth in memory usage with increasing rank, reflecting the storage cost of the correction subspace and the rank-dependent reduced linear algebra. By contrast, the wall-clock time exhibits a superlinear increase with rank, reflecting both the assembly of the reduced residual Jacobian and the direct solution of the reduced linear systems. These results illustrate the practical trade-offs between attainable accuracy and computational cost when selecting the subspace dimension.

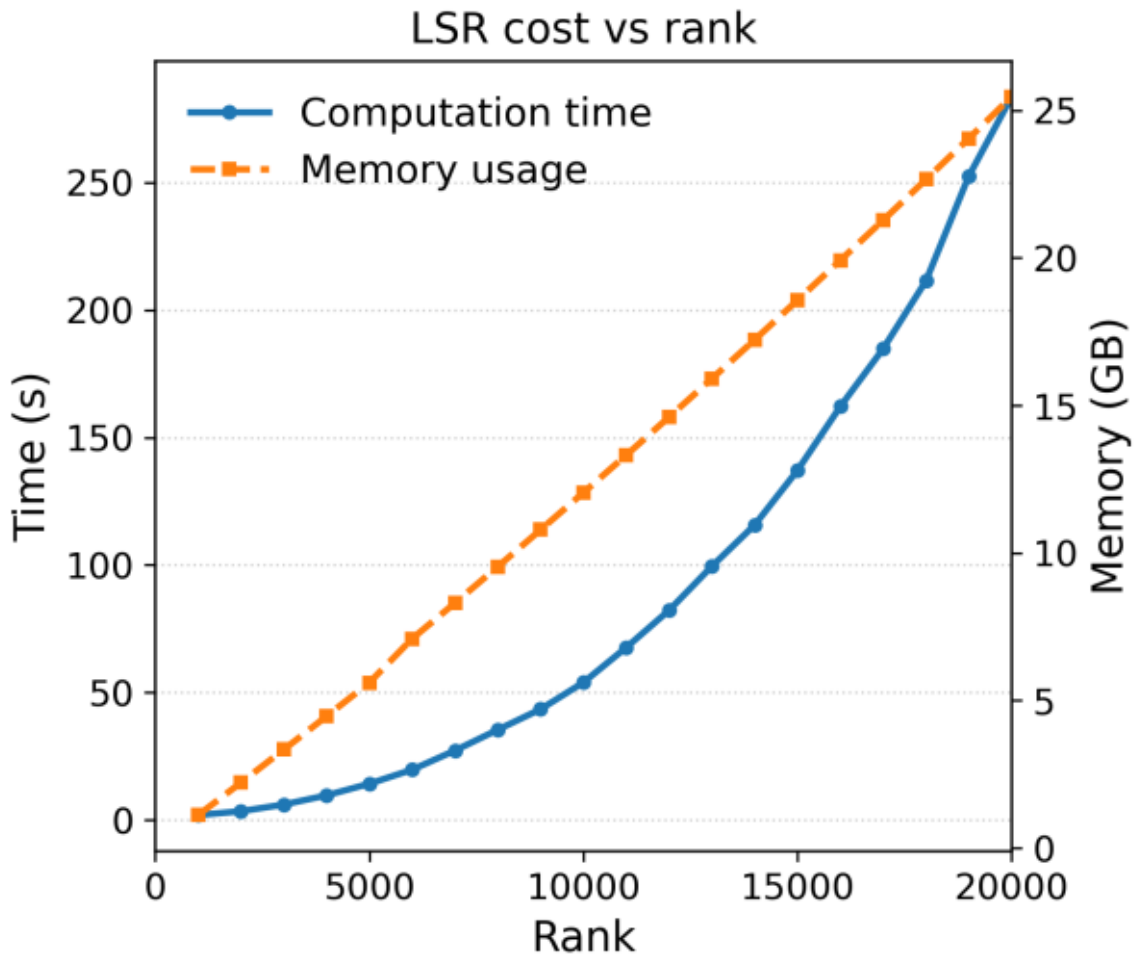

Supplementary Fig. 8. Memory usage and wall-clock time versus subspace rank.

## 2.3 Physics-informed operator fine-tuning

For the physics-informed operator fine-tuning experiments, all baseline models and training settings follow the configuration reported by Wang *et al.*[2]; detailed descriptions of the problem setup and training protocols can be found in their original work. Our implementation adopts the same model architectures and optimization strategies, with LSR introduced as an additional refinement step.

For LSR, the subspace rank is selected based on a coarse grid search by identifying the value that yields the lowest average loss over 30 instances. Using this criterion, the resulting ranks for the four equations are rank = 80, 200, 1000, and 800, respectively. Supplementary Fig. 9 presents results for two governing equations that complement those reported in the main text. Across all instances, LSR consistently improves both accuracy and stability relative to the baseline solutions.

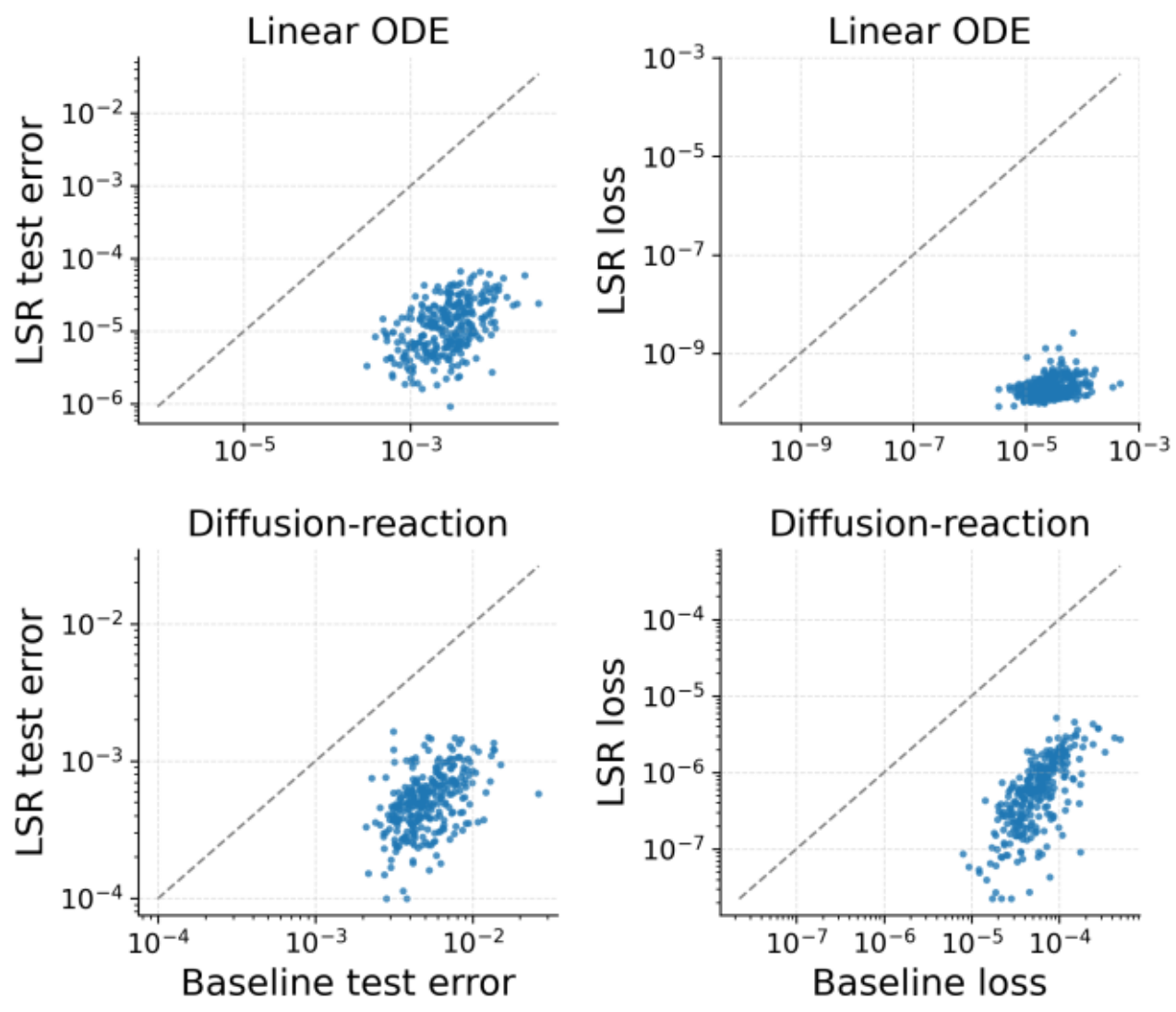

Supplementary Fig. 9. Baseline and LSR results for physics-informed operator fine-tuning.

To further characterize the numerical behavior of the correction subspaces in physics-informed operator fine-tuning, Supplementary Fig. 10 reports how the condition number of the reduced linear system and the norm of the corresponding LSR solution vary with the subspace dimension. The curves are obtained by averaging over 300 representative instances. As the subspace dimension increases, the reduced system becomes progressively more ill-conditioned, as reflected by the rapid growth of the condition number. In parallel, the norm of the reduced-space solution increases, indicating stronger amplification along poorly conditioned directions. Together, these diagnostics provide a quantitative visualization of the subspace-induced ill-conditioning that accompanies higher-dimensional refinements, and motivate the rank-selection discussed in the main text.

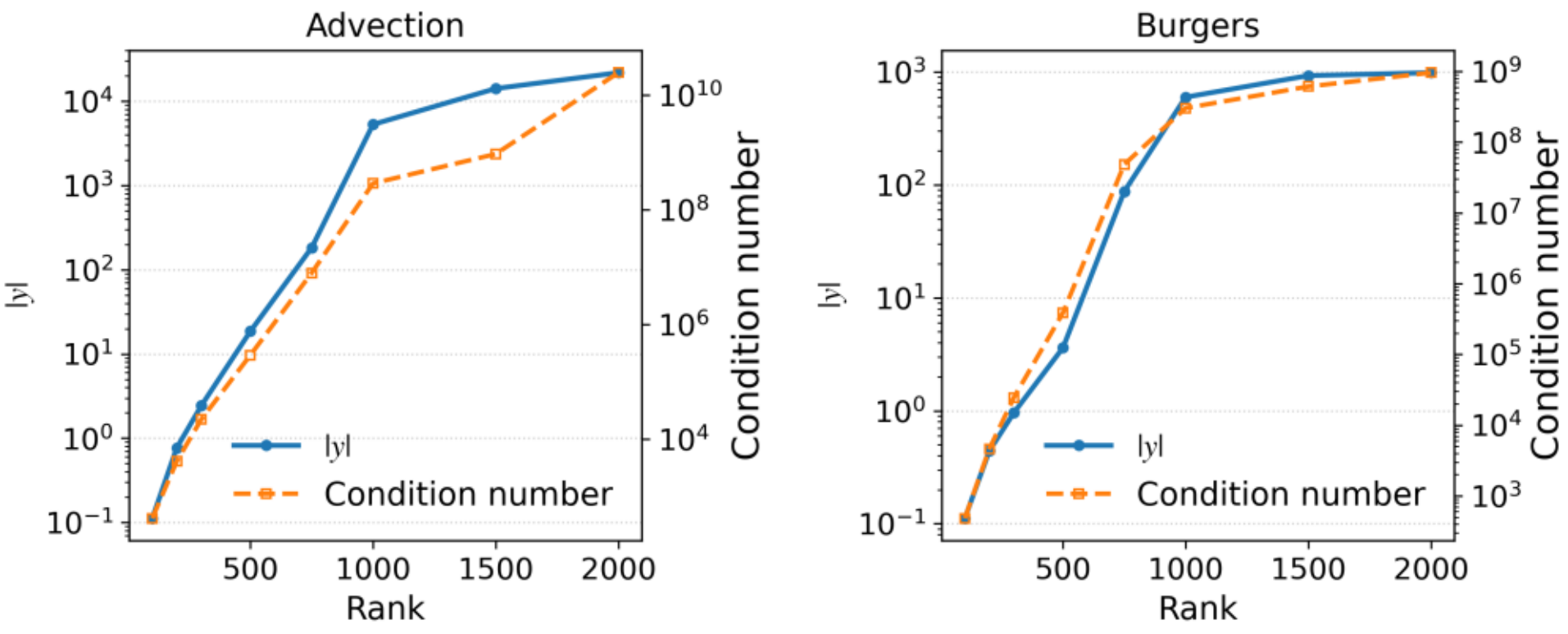

Supplementary Fig. 10. Subspace conditioning and solution norm in physics-informed operator fine-tuning.

## 2.4 Rank-controlled LSR for noisy inverse problems

The HInv experiment follows the heat inverse problem from the PINNacle benchmark[3]. Detailed definitions of the source term, boundary/initial conditions, observation sampling, and baseline configuration follow the original PINNacle paper and its official open-source implementation. Here we describe only the modifications introduced in this study. The original PINNacle benchmark does not use a validation split and considers a single noise level. In this work, we divide the noisy observations into 80% training data and 20% validation data. The validation set is used for two purposes: selecting the best baseline checkpoint during PINN training and selecting the LSR rank after post-training refinement. We also evaluate multiple noise levels to examine how both the baseline PINN and LSR behave as the observation noise changes.

To further support the rank-controlled LSR analysis in the HInv inverse problem, we report two additional diagnostic results. Supplementary Fig. 11 shows the convergence histories of the baseline PINN-LRA under different observation noise levels. Although reducing the noise level improves the final inversion accuracy, the baseline errors gradually saturate after training. In the low-noise regime, the final errors do not decrease proportionally with the observation noise, indicating that the dominant limitation is no longer only data quality, but also the optimization and approximation accuracy attainable by the trained PINN. Supplementary Fig. 12 reports the LSR loss as a function of rank for the same set of noise levels. The LSR loss decreases monotonically or nearly monotonically with increasing rank, showing that higher-rank subspaces provide stronger fitting capacity for the residual used in LSR. However, as shown in the main text, the clean temperature error and conductivity error may increase at large ranks under noisy observations. This contrast indicates that the high-rank degradation in the inverse problem is caused by overfitting or amplification of noisy observations, rather than by a failure of LSR to reduce its training objective. Therefore, validation-based rank selection is essential for balancing residual reduction and noise robustness.

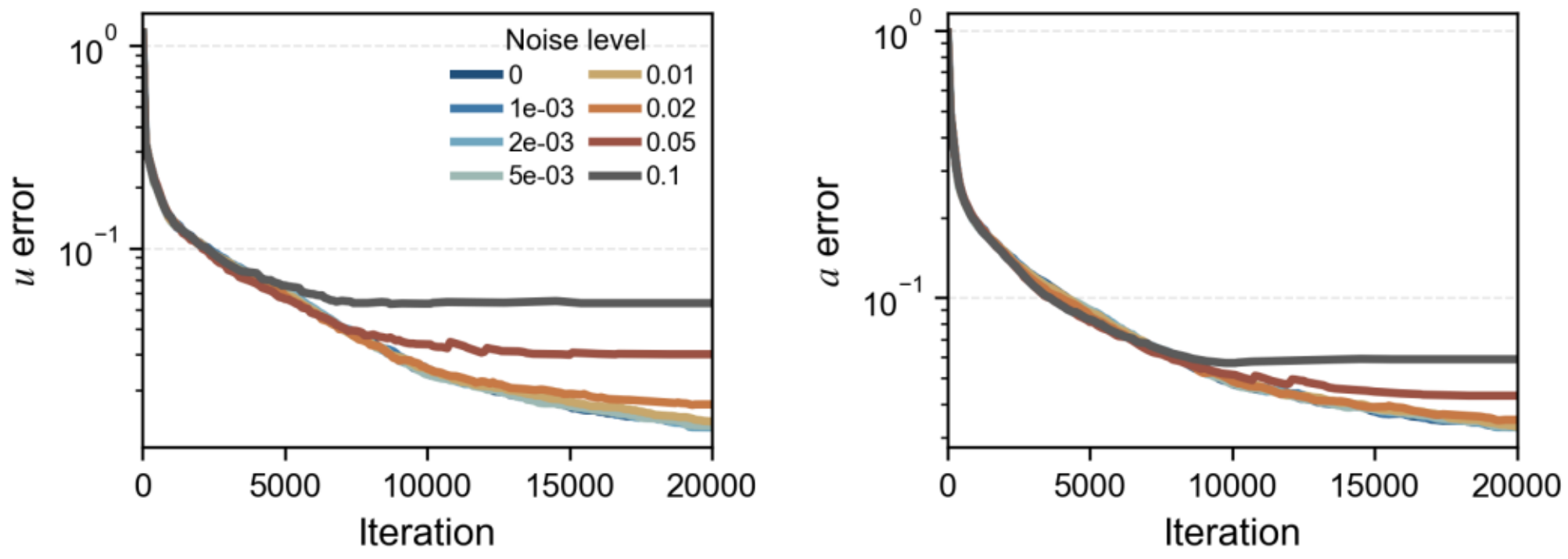

Supplementary Fig. 11. Baseline PINN convergence histories for the noisy HInv inverse problem for one random seed.

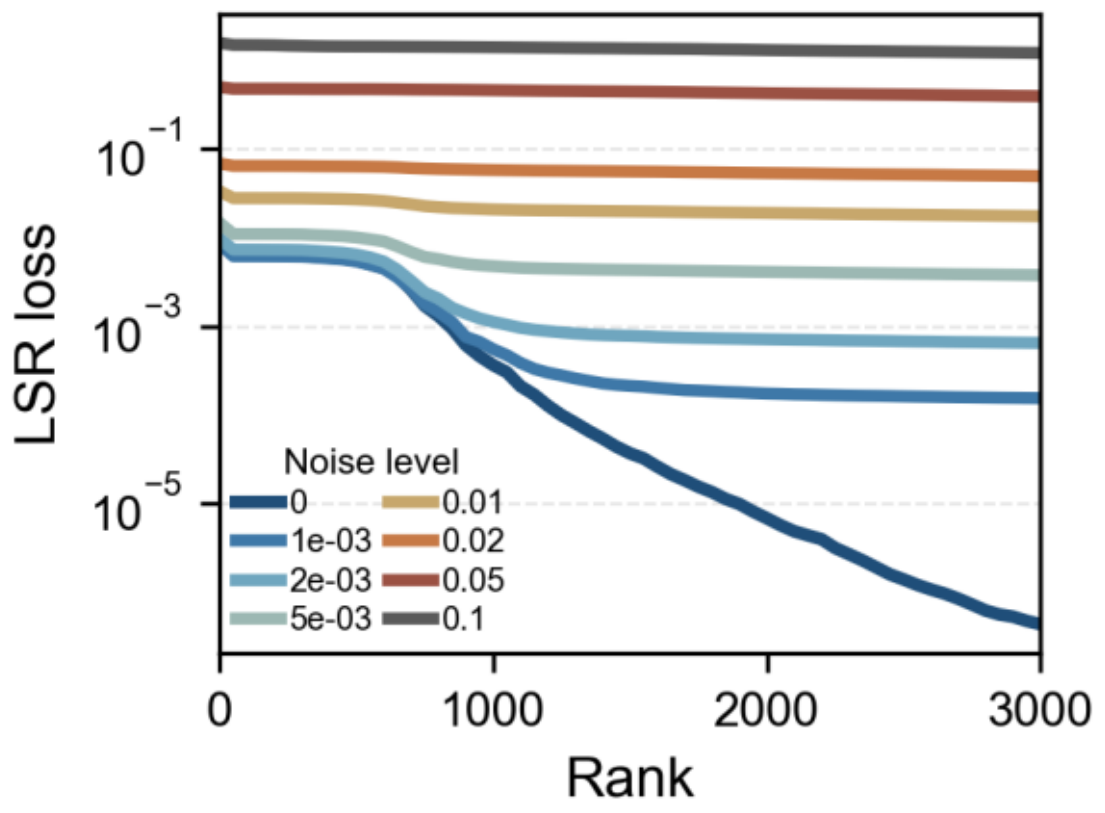

Supplementary Fig. 12. Rank-dependent LSR loss for the noisy HInv inverse problem for one random seed.

## 3. Comparison with finite-width empirical NTK regression

To further clarify the relationship between LSR and NTK-based linearized regression, we compare low-rank LSR with finite-width empirical NTK regression in the function approximation problem. The comparison is performed using the same trained base model as in the main function approximation experiment, so both methods start from the same trained state and the same local linearized model. Let $\mathbf{X}=\{x_i\}_{i=1}^{N}$ denote the training inputs, $\boldsymbol{y}$ the corresponding target values, and $\boldsymbol{q}_0(\mathbf{X})=\boldsymbol{q}(\mathbf{X};\boldsymbol{\theta}_0)$ the prediction of the trained base model.

In the empirical NTK formulation, the tangent kernel is defined as $\mathbf{K}(\mathbf{X},\mathbf{X})=\mathbf{J}(\mathbf{X})\mathbf{J}(\mathbf{X})^T$. The ridge-regularized empirical NTK kernel-regression correction is obtained by solving $(\mathbf{K}(\mathbf{X},\mathbf{X})+\lambda\mathbf{I})\boldsymbol{\alpha}=\boldsymbol{y}-\boldsymbol{q}_0(\mathbf{X})$, where $\lambda\geq 0$ is the ridge parameter. The unregularized empirical NTK regression corresponds to the limiting case $\lambda=0$, provided that the kernel system can be solved stably.

The corresponding prediction for a new input is $q_{NTK}(x)=q_0(x)+\boldsymbol{k}(x,\mathbf{X})\boldsymbol{\alpha}$, where $\boldsymbol{k}(x,\mathbf{X})=\mathbf{J}(x)\mathbf{J}(\mathbf{X})^T$ denotes the empirical NTK between the test input and the training inputs. For finite-width networks, this dual predictor can also be written in an equivalent primal correction form. Defining $\Delta\boldsymbol{\theta}_{NTK}=\mathbf{J}(\mathbf{X})^T\boldsymbol{\alpha}$, we obtain $q_{NTK}(x)=q_0(x)+\mathbf{J}(x)\Delta\boldsymbol{\theta}_{NTK}$.

Thus, after the full empirical NTK system has been solved, the finite-width NTK predictor can be evaluated in a primal form with a pointwise Jacobian-vector product, giving a prediction cost comparable to LSR. However, this does not remove the need to solve the full kernel system during refinement.

This distinction is important numerically. Full empirical NTK regression uses the entire tangent space, including directions associated with very small singular values. Moreover, because $\mathbf{K}=\mathbf{J}\mathbf{J}^T$, the nonzero eigenvalues of $\mathbf{K}$ are the squared singular values of $\mathbf{J}$, so the effective condition number satisfies $\kappa(\mathbf{K})=\kappa(\mathbf{J})^2$ over the nonzero spectrum. Therefore, the full empirical NTK system can be severely ill-conditioned in finite precision. In contrast, LSR restricts the correction to a rank-controlled correction subspace and solves the reduced least-squares problem directly, without forming the full empirical NTK matrix $\mathbf{JJ}^T$. This avoids the explicit kernel-system formulation that squares the Jacobian spectrum, while the rank truncation further excludes the most ill-conditioned high-order tangent directions.

Supplementary Table 1 compares rank-1000 LSR with full empirical NTK regression to clarify the distinction between tangent-space expressivity and numerical attainability. The trained baseline model has a train MSE of $5.25\times10^{-6}$ and a test MSE of $6.92\times10^{-6}$. Starting from the same base model, LSR reaches a train MSE of $4.70\times10^{-13}$ and a test MSE of $1.16\times10^{-12}$ in single precision, with further improvement observed in double precision. This result shows that a finite-rank Jacobian-induced response subspace already contains a highly effective correction that substantially improves the trained predictor. We then solve the corresponding full empirical NTK regression problem without ridge regularization. Although this full NTK formulation uses the complete tangent kernel rather than a reduced subspace, its single-precision result is numerically unstable, giving a train MSE of $1.12\times10^{-3}$ and a test MSE of $1.13\times10^{-3}$, both far worse than the baseline prediction. Double precision significantly improves the NTK result, reducing the train and test MSEs to $5.12\times10^{-11}$ and $1.51\times10^{-10}$, respectively. Nevertheless, even this double-precision NTK solve remains less accurate than the single-precision LSR result and incurs a substantially higher computational cost. This sharp precision dependence indicates that the high-accuracy solution associated with the full empirical NTK space is not necessarily numerically attainable through the full kernel solve under practical finite-precision computation. In contrast, LSR avoids forming and solving the full empirical kernel system $\mathbf{K}=\mathbf{JJ}^T$, whose spectrum inherits the squared singular-value structure of the Jacobian, and instead solves a rank-controlled reduced least-squares problem. This provides a more stable route to access useful hidden accuracy through the trained-state linearized predictor.

Supplementary Table 1. Computational cost and accuracy comparison between low-rank LSR and full empirical NTK regression. Note. NTK prediction time is reported for the standard dual kernel predictor. For finite-width networks, the solved NTK correction can also be represented in a primal form, giving a prediction cost comparable to LSR.

| Method | Precision | Solve time | Training MSE | Test MSE | Prediction time |
|---|---|---|---|---|---|
| LSR | fp32 | 2.02 s | 4.70E-13 | 1.16E-12 | 0.0012 s |
| LSR | fp64 | 13.95 s | 4.62E-15 | 2.38E-14 | 0.0087 s |
| NTK | fp32 | 2.71 s | 1.12E-03 | 1.13E-03 | 0.714 s |
| NTK | fp64 | 15.48 s | 5.12E-11 | 1.51E-10 | 16.81 s |

Beyond the refinement solve itself, the two formulations also differ in how the refined predictor is evaluated. Although the standard dual NTK predictor requires kernel evaluations against all training samples, the finite-width NTK solution can be converted into a primal correction form after the kernel system has been solved. For reference, the prediction time of the original neural network is $3.29\times10^{-4}$ s in single precision and $3.27\times10^{-3}$ s in double precision. Therefore, the LSR predictor, which requires one additional pointwise Jacobian-vector product, incurs only a total prediction cost approximately 2-4 times that of the original forward evaluation in this example.

We next examine the effect of ridge regularization on empirical NTK regression. As shown in Supplementary Table 2, increasing the ridge parameter stabilizes the single-precision NTK solve, and the single- and double-precision results eventually become nearly identical once the unstable directions of the full kernel system are sufficiently suppressed. However, this stabilization is achieved by shrinking the NTK correction. As the ridge parameter becomes large, the correction is almost completely damped, and the predictor approaches the original baseline model. Therefore, ridge regularization stabilizes the full NTK system only by trading away correction strength. By contrast, LSR controls the effective correction capacity through the subspace rank, retaining dominant response directions that provide useful refinement while avoiding the most unstable high-rank components. This comparison highlights a central distinction between ridge-regularized NTK regression and LSR: the former stabilizes a full kernel solve by globally suppressing the correction, whereas the latter seeks a numerically attainable low-rank correction that balances accuracy, stability, and computational cost.

Supplementary Table 2. Precision dependence and ridge-regularization effects in finite empirical NTK regression. The trained baseline model has a train MSE of $5.25\times10^{-6}$ and a test MSE of $6.92\times10^{-6}$.

| Ridge | $10^{-8}$ | $10^{-6}$ | $10^{-4}$ | $10^{-2}$ | $10^{-1}$ | $10^{1}$ | $10^{3}$ |
|---|---|---|---|---|---|---|---|
| fp32 train | 1.12E-03 | 1.12E-03 | 3.49E-03 | 9.05E-05 | 2.01E-06 | 5.16E-06 | 5.25E-06 |
| fp32 test | 1.13E-03 | 1.15E-03 | 3.53E-03 | 9.20E-05 | 3.01E-06 | 6.80E-06 | 6.92E-06 |
| fp64 train | 5.12E-11 | 2.04E-09 | 4.00E-08 | 3.92E-07 | 2.03E-06 | 5.16E-06 | 5.25E-06 |
| fp64 test | 1.51E-10 | 5.39E-09 | 8.47E-08 | 7.43E-07 | 2.97E-06 | 6.80E-06 | 6.92E-06 |

In this small example, the solve times of full empirical NTK regression and rank-1000 LSR are of the same order because the full kernel matrix is still tractable. The main difference is numerical stability: full NTK regression is much more sensitive to ill-conditioning. For larger datasets or high-dimensional residual problems, the empirical NTK formulation becomes substantially more expensive because it must handle an $N\times N$ Gram matrix, whereas LSR keeps the computation rank-controlled through a reduced correction subspace.

These results show that the main distinction between LSR and empirical NTK regression is not whether the tangent space is expressive. Full empirical NTK regression already uses the entire tangent space, but this full-space formulation can be strongly limited by numerical ill-conditioning. LSR instead identifies a numerically usable low-rank correction subspace and exploits it through a direct solve. Therefore, the LSR rank controls not only computational cost, but also the trade-off among expressive capacity, numerical stability, and practically attainable error reduction.

## 4. Linearized refinement versus nonlinear parameter update

This supplementary experiment is designed to clarify the interpretation and intended use of the LSR correction. In particular, we examine whether the correction direction obtained from the linearized least-squares problem can be used as a nonlinear parameter update.

Starting from the same trained network state used in the function approximation task, we consider the LSR correction direction $\Delta\boldsymbol{\theta}$ obtained by directly solving the reduced linearized least-squares problem. We then evaluate the nonlinear loss along the one-dimensional trajectory $\boldsymbol{\theta}(\alpha)=\boldsymbol{\theta}_0+\alpha\Delta\boldsymbol{\theta}$ for a range of step sizes $\alpha$. For comparison, we also evaluate the loss predicted by the linearized model at the same linearization point.

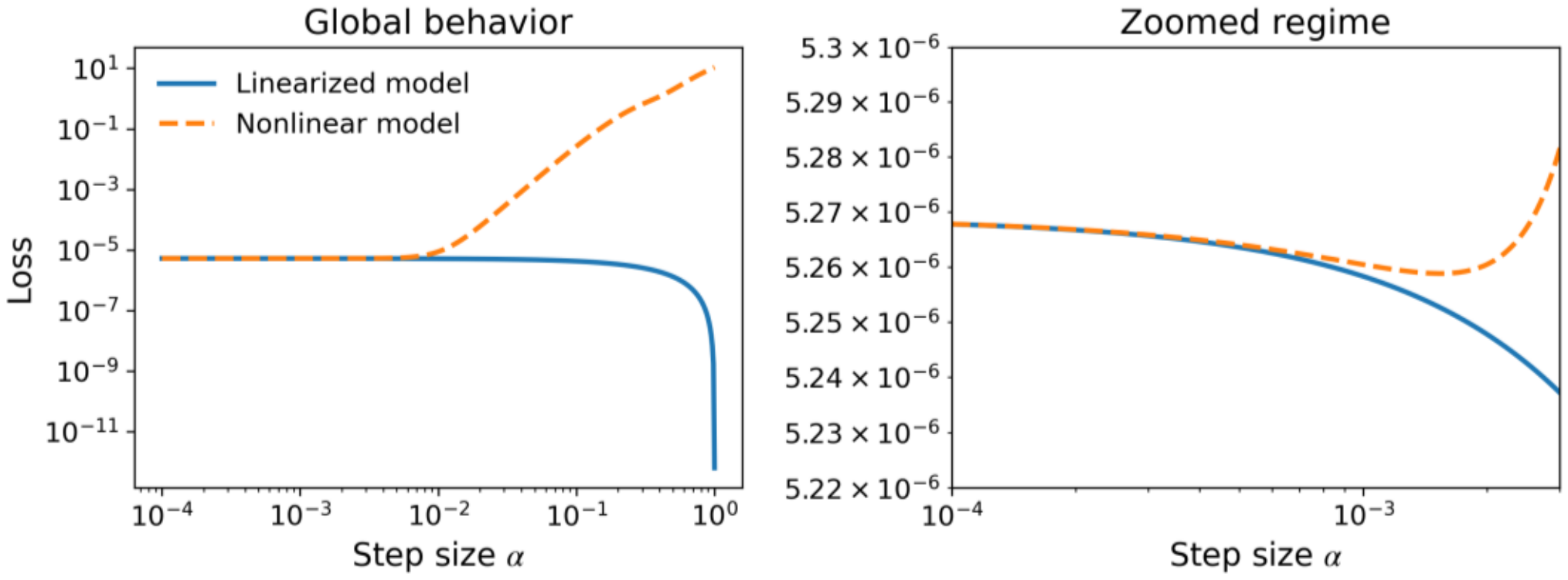

Supplementary Fig. 13. Nonlinear and linearized loss evaluated along the LSR correction direction.

The results are shown in Supplementary Fig. 13. While the loss predicted by the linearized model decreases monotonically along the LSR direction, the nonlinear loss exhibits a markedly different behavior. A slight reduction in the nonlinear loss can be observed only within a very narrow small-step regime, and the magnitude of this improvement is marginal compared with the reduction predicted by the linearized model. As the step size increases, the nonlinear loss rapidly deviates, reflecting the dominance of higher-order terms neglected by the local linear approximation.

These observations indicate that, although the linearized least-squares problem admits a solution with substantially lower residual, the corresponding correction direction does not constitute a reliable descent direction for the original nonlinear objective. Consequently, the LSR correction should not be interpreted as a nonlinear parameter update. Instead, LSR uses the linearized solution to define a refined predictor evaluated at the same linearization point, thereby improving accuracy without assuming global validity of the linear approximation.

## 5. Stationary-point behavior of LSR and a scalar illustrative example

### 5.1 Zero correction at first-order stationary points

LSR is derived by linearizing a residual minimization objective around a fixed trained state $\boldsymbol{\theta}_0$.

Consider the standard nonlinear least-squares problem

$$\min_{\boldsymbol{\theta}} \frac{1}{2}\|\boldsymbol{f}(\boldsymbol{\theta})\|_2^2, \tag{1}$$

where $\boldsymbol{f}(\boldsymbol{\theta})$ denotes the stacked residual vector and $\mathbf{G} = \partial \boldsymbol{f} / \partial \boldsymbol{\theta}$ is its Jacobian. The gradient of the objective is $\mathbf{G}^T \boldsymbol{f}$.

At a first-order stationary point $\boldsymbol{\theta}_0$, one has $\mathbf{G}^T \boldsymbol{f} = \boldsymbol{0}$.

One-shot LSR solves the linearized least-squares problem

$$\min_{\Delta\boldsymbol{\theta}} \frac{1}{2}\|\boldsymbol{f}(\boldsymbol{\theta}_0) + \mathbf{G}\Delta\boldsymbol{\theta}\|_2^2 . \tag{2}$$

The associated normal equations are

$$\mathbf{G}^T\mathbf{G}\Delta\boldsymbol{\theta} = -\mathbf{G}^T \boldsymbol{f}(\boldsymbol{\theta}_0). \tag{3}$$

Therefore, at a stationary point, the right-hand side vanishes and the linearized problem admits $\Delta\boldsymbol{\theta} = \boldsymbol{0}$ as a minimizer. In particular, if $\mathbf{G}$ has full column rank on the considered subspace, $\Delta\boldsymbol{\theta} = \boldsymbol{0}$ is the unique solution; if $\mathbf{G}$ is rank-deficient, any solution must satisfy $\mathbf{G}\Delta\boldsymbol{\theta} = \boldsymbol{0}$, which yields no change in the linearized prediction and hence no decrease in the linearized residual norm. In either case, LSR cannot produce a strictly improving correction at a first-order stationary point of the same residual objective.

This behavior is not a failure mode of LSR, but an expected consequence of stationarity: once the first-order optimality condition is satisfied, the local linear model provides no descent direction for the squared residual objective at $\boldsymbol{\theta}_0$.

The analysis above identifies a clear boundary case: LSR returns a zero correction when the underlying training has reached a first-order stationary point of the same residual objective. In practice, however, such stationary points are rarely attained under realistic computational budgets, even when the optimization problem is convex. As observed in our experiments, gradient-based training often stagnates at numerically induced plateaus and remains far from the solution that is accessible by a direct least-squares solve of the corresponding linearized problem.

These observations support the practical regime targeted by LSR: the method is designed to extract additional attainable accuracy at a fixed network state $\boldsymbol{\theta}_0$ when training has not fully satisfied the stationarity condition $\mathbf{G}^T \boldsymbol{f} = \boldsymbol{0}$. In such non-stationary yet stagnated states, the linearized least-squares correction computed by LSR is generally nonzero and can yield consistent refinement.

### 5.2 A scalar example illustrating the refined predictor

We next provide a one-dimensional example that visualizes how the linearized predictor family used by LSR can behave differently from the original nonlinear parametric mapping. Here the residual reduces to a scalar for visualization.

Let the scalar model output be

$$q(\theta) = (\theta - 1)^2 + 2, \quad (4)$$

and consider the infeasible target value $q^* = 1$, which corresponds to the desired target output in a supervised neural network fitting setting. Since $q(\theta) \geq 2$ for all $\theta$, the target cannot be matched by the original nonlinear parametric model.

Clearly, when the nonlinear parametric model attains its optimal performance (here at $\theta^* = 1$), linearization at this point yields no meaningful improvement. However, if the nonlinear model has not converged to a stationary point, i.e., $\theta \neq 1$, the linearized model at any such parameter value can render the target $q^* = 1$ attainable by choosing an appropriate $\Delta\theta$, as illustrated in Supplementary Fig. 14.

This simple example highlights two points relevant to LSR. First, the one-shot correction vanishes at stationary points because the linearized model admits no descent direction for the squared residual objective. Second, away from stationarity, the local linearized predictor family anchored at $\theta_0$ can produce corrections that outperform the original nonlinear model, indicating that the anchored linearized model may possess a richer effective expressive capacity than the underlying nonlinear parametric form.

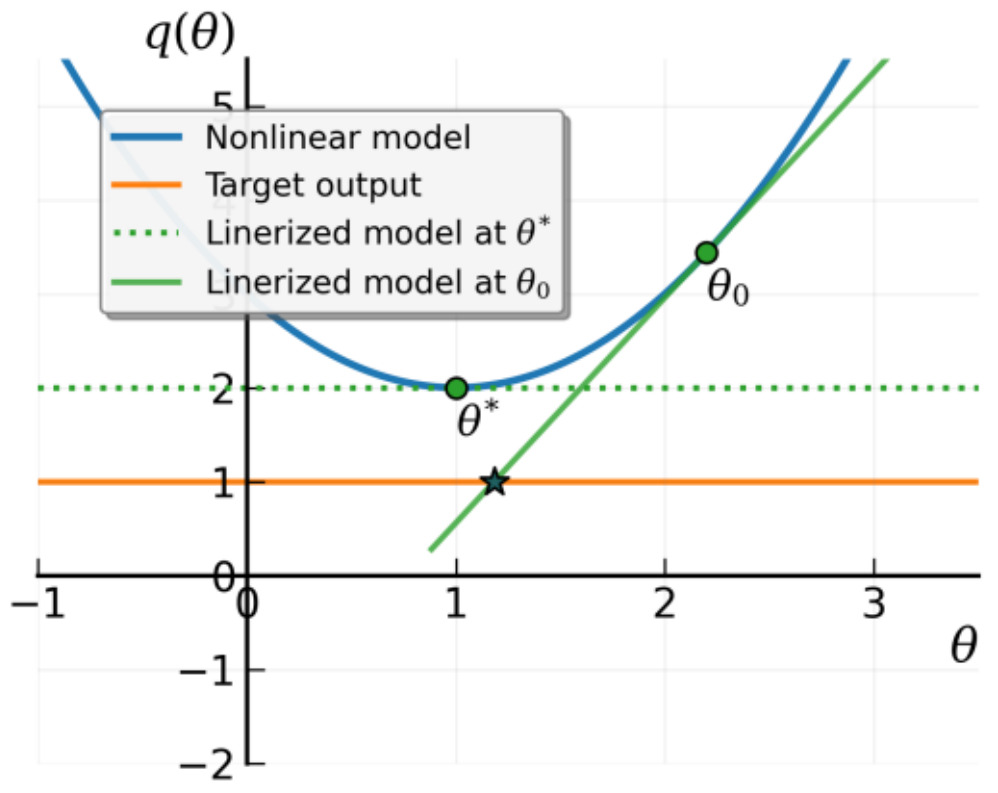

Supplementary Fig. 14. Scalar quadratic example illustrating linearized predictors in LSR.

## Supplementary References